%% file: main.tex
\pdfoutput=1

\documentclass[11pt]{article}

\usepackage[final]{coling}
\usepackage{booktabs}

\usepackage{times}
\usepackage{latexsym}

\usepackage[T1]{fontenc}

\usepackage[utf8]{inputenc}

\usepackage{microtype}

\usepackage{inconsolata}

\usepackage{graphicx}

\usepackage{amsmath}  
\usepackage{subcaption}
\usepackage{import}
\usepackage{multirow}
\usepackage{cleveref}
\crefname{equation}{Equation}{Equations}
\crefname{table}{Table}{Tables}
\crefname{figure}{Figure}{Figures}
\crefname{section}{Section}{Sections}
\crefname{appendix}{Appendix}{Appendices}

\usepackage{alltt}
\newcommand{\scorewithinterval}[2]{\begin{tabular}{@{}r@{}}#1\\[-1mm]\small{$\pm$#2}\end{tabular}}

\newcommand{\significantscorewithinterval}[2]{\phantom{\textsuperscript{*}}\begin{tabular}{@{}r@{}}#1\\[-1mm]\small{$\pm$#2}\end{tabular}\raisebox{0.5em}{\textsuperscript{*}}}

\newcommand{\highestscore}[1]{\textbf{#1}}

\newcommand{\modelName}[1]{\texttt{#1}}

\title{Self-Translate-Train: Enhancing Cross-Lingual Transfer\\of Large Language Models via Inherent Capability}

 \author{Ryokan Ri \and Shun Kiyono \and Sho Takase \\
         SB Intuitions \\
         \texttt{\{ryokan.ri,shun.kiyono,sho.takase\}@sbintuitions.co.jp}
 }

\begin{document}
\maketitle
\begin{abstract}
\input{contents/00-abstract}
\end{abstract}

\input{contents/01-introduction}
\input{contents/02-related_work}
\input{contents/03-framework}
\input{contents/04-experiments}
\input{contents/05-results}
\input{contents/06-conclusion}

\newpage
\input{contents/98-limitations}

\bibliography{contents/references}

\newpage
\onecolumn
\appendix

\input{contents/99-appendix}

\end{document}

%% file: contents/00-abstract.tex
Zero-shot cross-lingual transfer by fine-tuning multilingual pretrained models shows promise for low-resource languages, but often suffers from misalignment of internal representations between languages.
We hypothesize that even when the model cannot generalize across languages effectively in fine-tuning, it still captures cross-lingual correspondence useful for cross-lingual transfer.
We explore this hypothesis with Self-Translate-Train, a method that lets large language models (LLMs) to translate training data into the target language and fine-tunes the model on its own generated data.
By demonstrating that Self-Translate-Train outperforms zero-shot transfer, we encourage further exploration of better methods to elicit cross-lingual capabilities of LLMs.

%% file: contents/01-introduction.tex
\section{Introduction}
\label{sec:introduction}

Cross-lingual transfer is a technique to solve tasks in a target language by leveraging data from other languages~\citep{Pikuliak2021CrosslingualLF}.
A popular approach is to fine-tune a multilingual pre-trained model~\citep{conneau-etal-2020-unsupervised,xue-etal-2021-mt5,Scao2022BLOOMA1} on a source language data (e.g., English) and then apply it to the target language.
Capable models can solve tasks in the target language without seeing its data during fine-tuning, which is known as zero-shot cross-lingual transfer~\citep{artetxe-schwenk-2019-massively,chen-etal-2021-zero}.

However, zero-shot cross-lingual transfer is often suboptimal, with a performance gap between source and target languages.
One potential factor for this gap is representation misalignment between languages~\citep{artetxe-etal-2023-revisiting}: the model does not capture the correspondence between languages effectively.
We hypothesize that in some cases the model does capture the correspondence, but just cannot utilize it when fine-tuned only on the source language data; we need to elicit its cross-lingual capability in a different way.

To explore this hypothesis, we propose and investigate an approach called Self-Translate-Train, which leverages the powerful text generation capabilities of large language models (LLMs)~\citep{brownLanguageModelsAre2020,Touvron2023Llama2O}.
Instead of fine-tuning an LLM on source language data and hoping for generalization, Self-Translate-Train first let the model translate the training data into the target language, and fine-tunes the model on the data that the model itself generates.
This process does not only rely on cross-lingual generalization that happens during fine-tuning, but also on the cross-lingual correspondence that the model captures and can be elicited through translation.

This approach is inspired by the Translate-Train~\citep{conneau-etal-2018-xnli,pmlr-v119-hu20b,artetxe-etal-2023-revisiting} approach, a method to translate training data to the target language, often with an external machine translation system.
Notably, our findings reveal that the Translate-Train approach remains effective even without external resources.

\begin{figure}[t!]
    \includegraphics[width=\columnwidth]{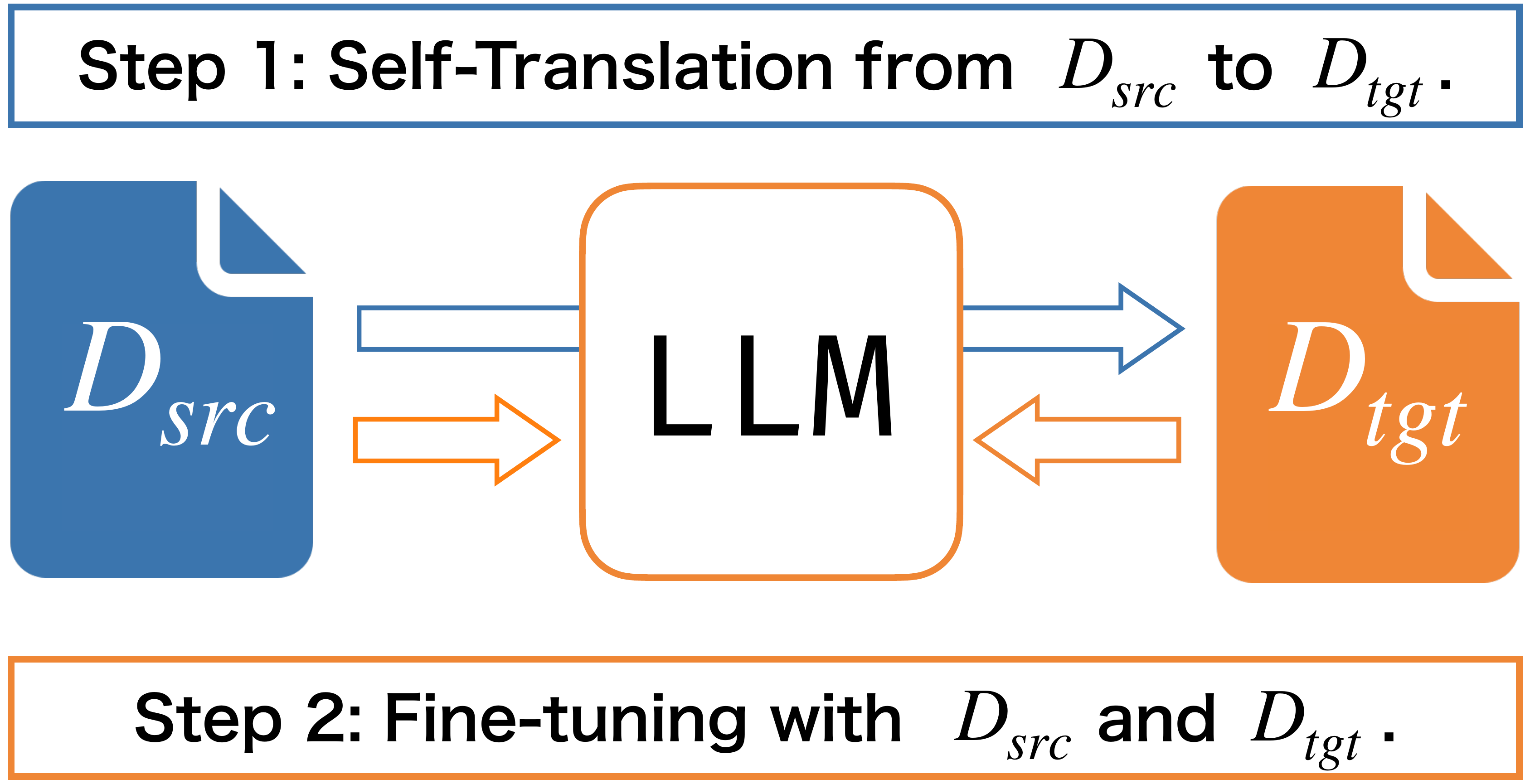}
    \caption{An overview of Self-Translate-Train.
    }
    \label{fig:overview}
    \vspace{-10pt}
\end{figure}

We validate the effectiveness of Self-Translate-Train with three tasks (math, QA, NLI) across four target languages (de, ru, th, zh).
The results show that better cross-lingual performance can be achieved by correctly eliciting the model's translation capability, which encourages further exploration of a better method for cross-lingual transfer.

%% file: contents/02-related_work.tex
\section{Related Work}
\label{sec:related_work}

\vspace{-2pt}
\subsection{Cross-lingual Transfer Learning}
\label{subsec:cross-lingual_transfer_learning}
\vspace{-2pt}

There are two main approaches to transfer task knowledge across languages: model transfer and data transfer~\citep{Pikuliak2021CrosslingualLF}.

\noindent\textbf{Model transfer} leverages multilingual pre-trained models.
As the model is expected to capture the commonality between languages, when fine-tuned on source language data, it can generalize to solve the task in other languages~\citep{pires-etal-2019-multilingual,mulcaire-etal-2019-polyglot,conneau-etal-2020-unsupervised}.

\noindent\textbf{Data transfer} translates the source language data into the target language.
In the Translate-Test approach, models are assumed to be specialized in the source language and the task inputs are translated into this language at test time~\citep{conneau-etal-2018-xnli,etxaniz-etal-2024-multilingual}.
The Translate-Train approach translates the training data, and the trained model makes predictions in the target language~\citep{conneau-etal-2018-xnli}.
Data transfer is often more effective than model transfer~\citep{pmlr-v119-hu20b,artetxe-etal-2023-revisiting,ebing-glavas-2024-translate}.

Our approach, Self-Translate-Train, directly translates the training data as in data transfer, but also leverages the cross-lingual capabilities of the model itself as in model transfer.

\vspace{-2pt}
\subsection{Self-Improvement of LLMs}
\label{subsec:self-improvment_of_llms}
\vspace{-2pt}

LLMs have remarkable text generation capabilities, which has been leveraged to generate training data for various purposes~\citep{li-etal-2023-synthetic,Lee2024GeckoVT}.
The generated data can be used to further specialize the LLM itself for downstream applications, without requiring an extensive collection of additional data.
This process can be viewed as a form of self-improvement~\citep{Bai2022ConstitutionalAH,huang-etal-2023-large,DBLP:conf/nips/SunSZZCCYG23,DBLP:journals/corr/abs-2308-06259}.

Self-Translate-Train is also a self-improvement approach to specialize the LLM to a target language by translating the source language data.

%% file: contents/03-framework.tex
\vspace{-2pt}
\section{Self-Translate-Train}
\label{sec:method}
\vspace{-2pt}

Our framework focuses on fine-tuning LLMs on a small amount of data for a specific task.
Let the training corpus in a source language, say English, be $\mathcal{D}_{\text{src}} = \{(\mathbf{x}_{\text{src}}^i, \mathbf{y}_{\text{src}}^i)\}_{i=1}^N$, where $\mathbf{x}$ is the input and $\mathbf{y}$ is the output.
In a typical cross-lingual transfer setting, the model is fine-tuned only on $\mathcal{D}_{\text{src}}$ and expected to generalize to target languages.

\subsection*{Translated Synthetic Data}

Given the LLM's translation capability, we can let it translate the training corpus into a synthetic corpus in the target language $\mathcal{D}_{\text{tgt}}$.
The synthetic data can be added to $\mathcal{D}_{\text{src}}$ to achieve a better generalization to the target language.

The translation can be performed in various ways depending on the model's capabilities or available resources.
In this paper, we experiment with the few-shot prompting technique (\cref{subsec:synthetic_data_generation}).

\subsection*{Code-switched Synthetic Data}

The generated data has an interesting aspect: each synthetic instance has a corresponding instance in the original dataset with the same semantics.
We can exploit this to further synthesize data by generating code-switched instances where the input and output are in different languages.

We pair the original and translated instances to construct $\mathcal{D}_{\text{cs}} = \{(\mathbf{x}_{\text{src}}^i, \mathbf{y}_{\text{tgt}}^i)\}_{i=1}^N \bigcup \{(\mathbf{x}_{\text{tgt}}^i, \mathbf{y}_{\text{src}}^i)\}_{i=1}^N$.
When the task output is natural language, we manually translate the prompt ``Please answer in \{\{ tgt \}\}.'' into the target language, and add it to the input $\mathbf{x}$.

%% file: contents/04-experiments.tex
\section{Experimental Setups}
\label{sec:experimental_setups}

To verify the effectiveness of Self-Translate-Train, we conduct extensive experiments on multiple tasks and languages.

\subsection{Task and Datasets}
\label{subsec:datasets}

We present a list of datasets for experiments in Table~\ref{tab:list-of-datasets}.
For each task, an English dataset is used for training and a multilingual dataset for evaluation.
To make the computational cost feasible, we use a 10,000-sample subset of the training data for SQuAD and MultiNLI.

\begin{table}[th]
    \small
    \centering
    \begin{tabular}{@{}lll@{}}
    \toprule
    Task                     & Training & Evaluation \\ \midrule
    QA       & \parbox[t]{2cm}{SQuAD\vspace{-2pt} \\ {\tiny\citep{rajpurkar-etal-2016-squad}}} & \parbox[t]{2cm}{XQuAD\vspace{-2pt} \\ {\tiny\citep{artetxe-etal-2020-cross}}} \\
    Classification & \parbox[t]{2cm}{MultiNLI\vspace{-2pt} \\ {\tiny\citep{williams-etal-2018-broad}}} & \parbox[t]{2cm}{XNLI\vspace{-2pt} \\ {\tiny\citep{conneau-etal-2018-xnli}}} \\
    Math   & \parbox[t]{2cm}{GSM8k\vspace{-2pt} \\ {\tiny\citep{Cobbe2021TrainingVT}}} & \parbox[t]{2cm}{MGSM\vspace{-2pt} \\ {\tiny\citep{DBLP:conf/iclr/ShiSF0SVCTRZ0W23}}} \\ \bottomrule
    \end{tabular}
    \caption{List of datasets for experiments. The details are described in \cref{appendix:datasets}.}
    \label{tab:list-of-datasets}
    \vspace{-3pt}
\end{table}

\vspace{-3pt}
\subsection{Languages}
\label{subsec:languages}
\vspace{-3pt}

We conducted evaluation on four languages: German (de), Russian (ru), Thai (th), and Chinese (zh).
German is a Germanic language, which is phylogenetically close to English and expected to show better cross-lingual transfer, while Russian, Thai, and Chinese are from different language families.
In particular, Thai is a low-resource language with a different script from English, which is expected to be more challenging for cross-lingual transfer.

\vspace{-3pt}
\subsection{Language Models}
\label{subsec:language_models}
\vspace{-3pt}

Our main experiments use \modelName{Llama2-7B}~\citep{Touvron2023Llama2O}, a public LLM.
Although 90\% of its pretraining corpus is English, the model has a multilingual capability (e.g., \cref{tab:bleu_mgsm}) from the remaining fraction of multilingual data.

\vspace{-3pt}
\subsection{Synthetic Data Generation}
\label{subsec:synthetic_data_generation}
\vspace{-3pt}

Recent LLMs are known to exhibit a translation capability without much task-specific data~\cite{briakou-etal-2023-searching}.
In our experiments, we elicit the translation capability of the LLMs via few-shot in-context learning~\citep{brownLanguageModelsAre2020}.

To construct few-shot translation samples, we sample eight pairs from the train or validation splits of the multilingual datasets, where instances across languages form parallel data.
The translation was performed for each field individually, e.g., for GSM8k, we translated the question and answer separately.
The prompt template simply alternates the source and target text prepended with the language tag (\cref{appendix:prompt_format}).

An important step to ensure the quality of the synthetic data is to filter out the low-quality data (the details in \cref{appendix:data_filtering}).
To remove under- or over-translation~\citep{tu-etal-2016-modeling}, we filter out texts with an extreme source-target length ratio.
Also, to address the repetition problem~\citep{DBLP:conf/iclr/HoltzmanBDFC20}, we set the max number of tokens for generation and filter out the translation that does not end with the EOS token.
With the translations from \modelName{Llama2-7B}, this process removes around 10\% of the data for most languages and around 50\% for Thai due to the model's limited generation quality.

To provide the sense of the translation quality, we report the BLEU score~\citep{papineni-etal-2002-bleu} measured by the parallel data constructed from questions in the MGSM test set in \cref{tab:bleu_mgsm}.
Overall, the translation quality is sufficiently high except for Thai.
As we will see in \cref{subsec:main_results}, this poses a challenge for cross-lingual transfer to Thai.

\begin{table}[h]
\centering
    \setlength{\tabcolsep}{0.35em}
    \scalebox{0.9}{
    \begin{tabular}{lcccccc}
    \toprule
    \textbf{Model}  & \textbf{de}   & \textbf{ru} & \textbf{th} & \textbf{zh} \\ \midrule
    \modelName{Llama2-7B}       & 37.1          & 27.2        & 1.9         & 29.4        \\
    \bottomrule
    \end{tabular}
    }
\caption{BLEU scores from the MGSM test set with 8-shot trasnlation. The details of the configuration is described in \cref{appendix:translation_quality}.}
\vspace{-7pt}
\label{tab:bleu_mgsm}
\end{table}
\vspace{-7pt}

\subsection{Fine-tuning}
\label{subsec:fine_tuning}

All the tasks are cast as text generation tasks, where the LLM is given the inputs as a prompt and generate the answer.
Fine-tuning is conducted with causal language modeling loss, computed only for output tokens.
We use LoRA~\citep{DBLP:conf/iclr/HuSWALWWC22}, a parameter-efficient tuning technique, to reduce computational cost.

We use AdamW~\citep{DBLP:conf/iclr/LoshchilovH19} and the cosine learning rate schedule for optimization, training with a batch size of 64 for 1,000 steps.
For each setting, we conduct six runs with two learning rates (5e-5 and 3e-4) and different random seeds, reporting summarization statistics of the top four runs based on validation accuracy to remove runs with optimization failure.
See \cref{appendix:hyperparameters} for other hyperparameters.

%% file: contents/05-results.tex
\vspace{-3pt}
\section{Results}
\label{sec:results}
\vspace{-3pt}

\subsection{Main Results}
\label{subsec:main_results}

\input{contents/main_results_table.tex}

\begin{figure*}[!ht]
\begin{subfigure}{0.33\textwidth}
    \includegraphics[width=\linewidth]{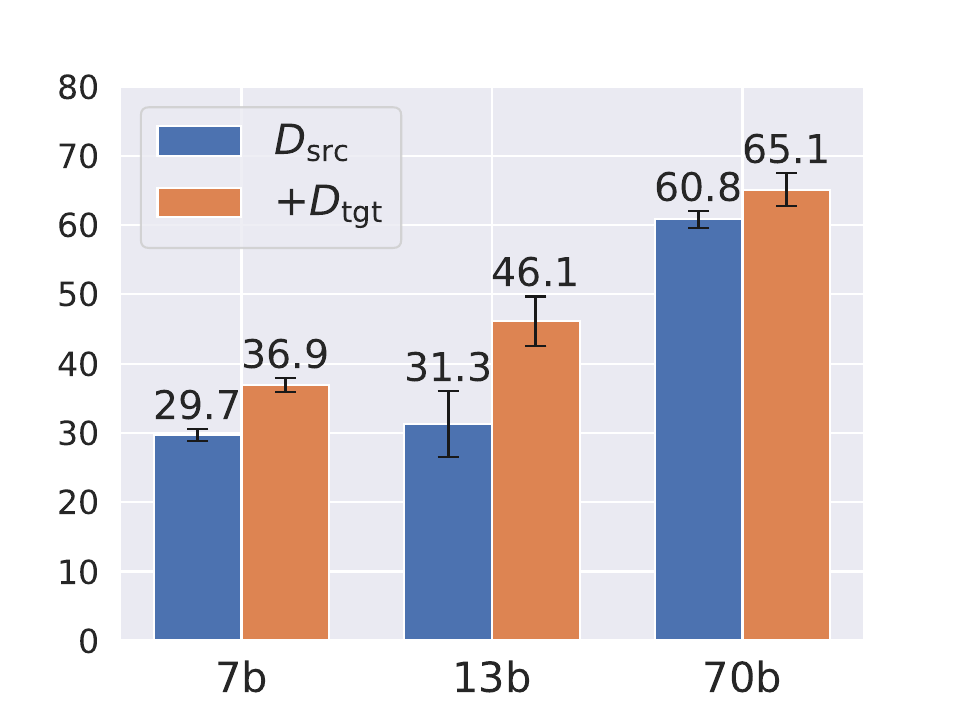}
    \caption{de}
\end{subfigure}%
\begin{subfigure}{0.33\textwidth}
    \includegraphics[width=\linewidth]{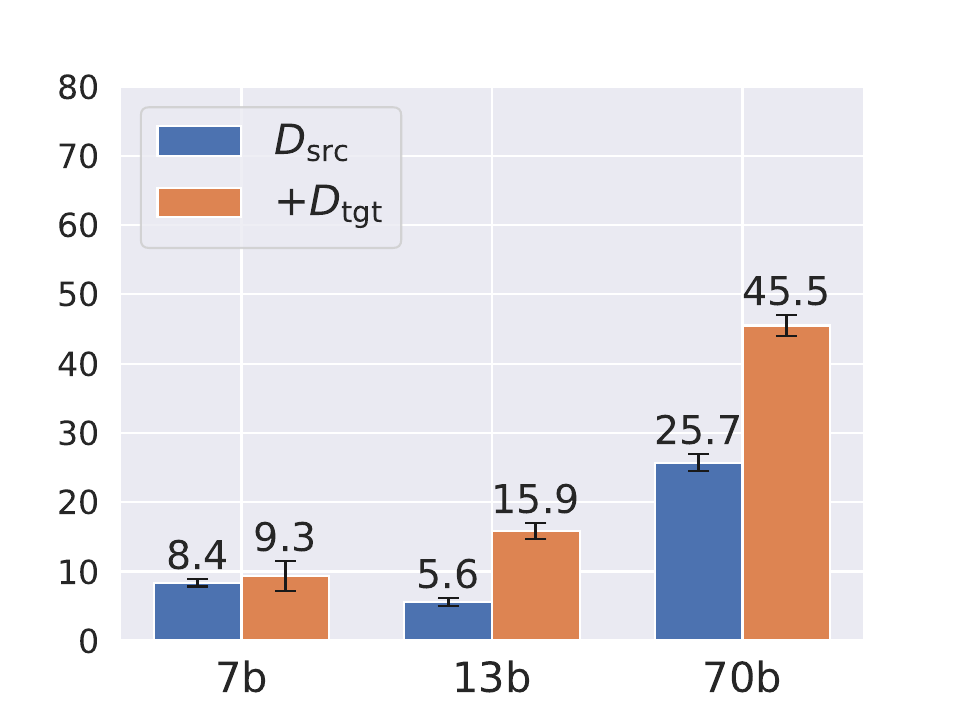}
    \caption{th}
\end{subfigure}%
\begin{subfigure}{0.33\textwidth}
    \includegraphics[width=\linewidth]{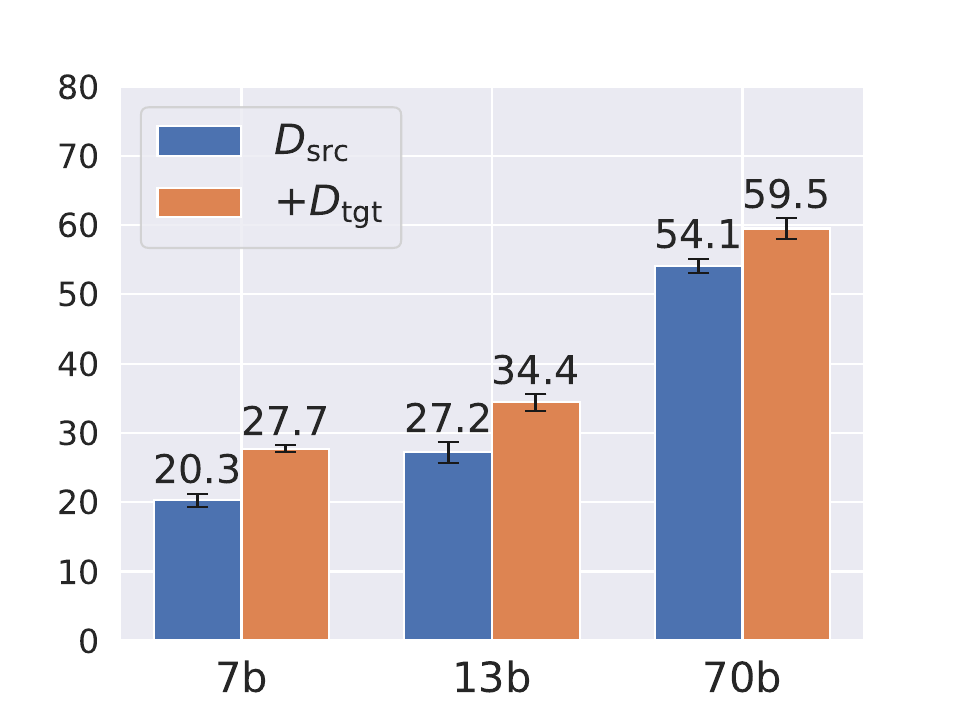}
    \caption{zh}
\end{subfigure}%
    \caption{Accuracy in the MGSM dataset with different model sizes of \modelName{Llama2}.}
    \label{fig:compare_sizes}
    \vspace{-5pt}
\end{figure*}

As a baseline, we perform few-shot learning with the target language data with the pre-trained model (\texttt{FewShot}).
We also fine-tune the LLM with the source language dataset $\mathcal{D}_{\text{src}}$.
To ensure equal data resources, we augment $\mathcal{D}_{\text{src}}$ with the eight target language samples used for few-shot translation (\cref{subsec:synthetic_data_generation}).
We then compare these baselines with the models fine-tuned on the data generated from Self-Translate-Train ($\mathcal{D}_{\text{src}}$ and $\mathcal{D}_{\text{cs}}$) in \cref{tab:main_results}\footnote{For experiments with other approaches, including translate-test and translation with external systems, see \cref{appendix:additional_experiments}.}.

Results show that fine-tuning on English ($\mathcal{D}_{\text{src}}$) is more effective than few-shot learning in the target language (\texttt{FewShot}).
This confirms that cross-lingual transfer from English is non-trivial and effective in this setting.

Notably, Self-Translate-Train is indeed an effective method; $+\mathcal{D}_{\text{tgt}}$ almost consistently outperforms the baseline $\mathcal{D}_{\text{src}}$.
The only exception is Thai (th), where there is no significant improvement.
This is likely due to the low translation quality of the model in Thai (\cref{tab:bleu_mgsm}).

However, the effectiveness of code-switching dataset is limited.
When we add $\mathcal{D}_{\text{cs}}$ to $\mathcal{D}_{\text{tgt}}$, there is no significant improvement from adding $\mathcal{D}_{\text{tgt}}$ only ($p < 0.05$ in Welch's t-test).
This indicates that the code-switching data does not provide additional information for the model to generalize in the task.

\subsection{Does the model size matter?}
\label{subsec:model_size}

The size of the LLM can influence both its ability to generalize across languages and the quality of its translations, which in turn may impact the effectiveness of Self-Translate-Train.
We compare the performance of \modelName{Llama2} with different sizes, i.e., 7B, 13B, and 65B, on the math task in de, th, and zh.
For each model size, we generate $\mathcal{D}_{\text{tgt}}$ with the corresponding model and then fine-tune itself.

\cref{fig:compare_sizes} shows that the larger model size generally tends to perform better, and the improvement from Self-Translate-Train remains consistent across different model sizes.
In Thai (th), we did not observe a significant improvement in the 7B model, but do in the larger models (13B and 65B), likely due to their better translation quality.
The 7B model has a low Thai translation BLEU score of 1.9 (\cref{tab:bleu_mgsm}), while the 13B and 65B models have BLEU scores of 5.1 and 12.0, respectively.

The improvement in Thai (th) with the 70B model is the most significant (+19.8 average points).
This implies that Self-Translate-Train is particularly effective when the model struggles with generalizing across the source and target languages but can still generate their reasonable translations.

%% file: contents/main_results_table.tex
\begin{table*}[h!]
\centering
    \setlength{\tabcolsep}{0.40em}
    \scalebox{0.90}{
    \begin{tabular}{l|cccc|cccc|cccc}
\toprule
 & \multicolumn{4}{|c|}{MGSM} & \multicolumn{4}{|c|}{XQuAD} & \multicolumn{4}{|c}{XNLI} \\
 & de & ru & th & zh & de & ru & th & zh & de & ru & th & zh \\ \midrule

\texttt{FewShot} & \scorewithinterval{11.2}{1.4} & \scorewithinterval{11.9}{1.5} & \scorewithinterval{3.3}{0.9} & \scorewithinterval{10.3}{0.2} & \scorewithinterval{44.8}{0.7} & \scorewithinterval{30.9}{0.3} & \scorewithinterval{24.7}{0.4} & \scorewithinterval{51.5}{0.7} & \scorewithinterval{41.8}{0.1} & \scorewithinterval{39.0}{0.2} & \scorewithinterval{34.2}{0.0} & \scorewithinterval{38.5}{0.3} \\ \hline
$\mathcal{D}_{\text{src}}$ & \scorewithinterval{30.1}{0.4} & \scorewithinterval{25.0}{0.7} & \scorewithinterval{8.1}{0.7} & \scorewithinterval{21.1}{1.7} & \scorewithinterval{60.3}{0.8} & \scorewithinterval{49.3}{0.4} & \scorewithinterval{34.5}{1.0} & \scorewithinterval{66.3}{0.8} & \scorewithinterval{79.7}{0.4} & \scorewithinterval{76.9}{0.1} & \scorewithinterval{53.7}{0.9} & \scorewithinterval{74.1}{0.2} \\
\hline
$+\mathcal{D}_{\text{tgt}}$ & \significantscorewithinterval{\highestscore{36.4}}{1.3} & \significantscorewithinterval{34.0}{1.7} & \scorewithinterval{7.7}{3.4} & \significantscorewithinterval{27.1}{1.2} & \scorewithinterval{61.7}{0.9} & \significantscorewithinterval{57.8}{0.8} & \significantscorewithinterval{\highestscore{46.4}}{1.3} & \significantscorewithinterval{\highestscore{77.7}}{0.4} & \significantscorewithinterval{\highestscore{81.6}}{0.7} & \significantscorewithinterval{\highestscore{78.5}}{0.6} & \scorewithinterval{56.3}{1.5} & \significantscorewithinterval{\highestscore{78.5}}{0.3} \\
$+\mathcal{D}_{\text{tgt}} + \mathcal{D}_{\text{cs}}$ & \significantscorewithinterval{35.9}{1.3} & \significantscorewithinterval{\highestscore{34.5}}{2.4} & \scorewithinterval{10.4}{1.6} & \significantscorewithinterval{\highestscore{28.8}}{1.9} & \significantscorewithinterval{62.2}{0.9} & \significantscorewithinterval{\highestscore{58.0}}{0.6} & \significantscorewithinterval{46.2}{1.7} & \significantscorewithinterval{77.3}{0.7} & \significantscorewithinterval{81.6}{0.5} & \scorewithinterval{78.4}{1.2} & \significantscorewithinterval{\highestscore{58.9}}{1.5} & \significantscorewithinterval{77.6}{0.6} \\
\bottomrule
\end{tabular}
    }
\caption{Results on multilingual evaluation datasets.
    Scores are marked with $^*$ if its improvement is statistically significant ($p < 0.05$ in Welch's t-test) compared to the baseline $\mathcal{D}_{\text{src}}$.
    The significant and highest score in each column is marked in bold. The FewShot baseline uses 8 samples, except for XQuAD where model token limits (4096) require fewer: 2 for Russian and 1 for Thai.
}
\vspace{-5pt}
\label{tab:main_results}
\end{table*}

%% file: contents/06-conclusion.tex
\section{Discussion and Conclusion}
\label{sec:conclusion}

We explored Self-Translate-Train, a method to improve cross-lingual transfer performance by generating synthetic training data in the target language.
We validated its effectiveness on various tasks and languages, demonstrating substantial performance gains over baselines with the same data budget.

This result aligns with \citet{etxaniz-etal-2024-multilingual}, who showed that the performance of few-shot learning can be improved by self-translating the task inputs.
Our work demonstrates that self-translation by an LLM is also effective when applied to task-specific training data for fine-tuning.

These results suggest that we can improve the cross-lingual transfer performance of an LLM without any additional data, but by investing more computation to generate synthetic data.
This finding is consistent with the recent trend of self-improvement methods for LLMs~\citep[e.g.,][]{huang-etal-2023-large}.

Future work could explore synthetic data generation for pre-training to improve cross-lingual capabilities of LLMs without intensive data collection.
We hope that this work encourages further exploration of how to better elicit models' cross-lingual capabilities.

%% file: contents/98-limitations.tex
\section{Limitations}
\label{sec:limitation}

Our experiments are conducted on a modern type of LLM, an autoregressive Transformer decoder, and centered around the Llama2 model families~\citep{Touvron2023Llama2O}.
Although we further validate our method in \cref{appendix:results_qwen}, the effective of our proposed method is uncertain when applied to other types of LLMs developed in the future.

Our method is based on the assumption that the model can generate reasonable translations in the target language.
However, for the low-resource target language or the datasets with out-of-domain text for the model, the translation quality can be not satisfactory for the Translate-Train approach~\citep{ebrahimi-wense-2024-zero}.

Finally, when the task requires generating long and natural text, the quality of the generated translation matters more.
If the translation quality is low, the model outputs may degrade due to translation errors or unnaturalness.
The application of our method on more challenging tasks requires further technical considerations.

%% file: contents/99-appendix.tex
\section{The Details of Experimental Setups}
\label{appendix:experimental_setups}

\subsection{Tasks and Datasets}
\label{appendix:datasets}

We provide the details of the datasets introduced in \cref{subsec:datasets}.

\subsubsection{Question Answering (QA)}
SQuAD~\citep{rajpurkar-etal-2016-squad} is an English QA dataset created from Wikipedia articles as training data.
Given a question and a passage, the task is to extract the answer from the passage.
Evaluation is conducted with XQuAD~\citep{artetxe-etal-2020-cross}, which consists of translation of SQuAD into multiple languages.

\begin{figure}[h]
  \begin{center}
  \begin{tabular}{p{0.95\columnwidth}}
      \toprule
\begin{alltt}
Context: Architecturally, the school has a Catholic character. Atop  the
Main Building's gold dome is a golden statue of the Virgin Mary. Immediately
in front of the Main Building and facing it, is a copper statue of Christ
with arms upraised with the legend "Venite Ad Me Omnes". Next to the Main
Building is the Basilica of the Sacred Heart. Immediately behind the basilica
is the Grotto, a Marian place of prayer and reflection. It is a replica of the
grotto at Lourdes, France where the Virgin Mary reputedly appeared to Saint
Bernadette Soubirous in 1858. At the end of the main drive (and in a direct line
that connects through 3 statues and the Gold Dome), is a simple, modern stone
statue of Mary.
Question: To whom did the Virgin Mary allegedly appear in 1858 in Lourdes France?

\end{alltt}
\\\midrule
\begin{alltt}
Saint Bernadette Soubirous

\end{alltt}
    \\ \bottomrule
  \end{tabular}
  \end{center}
  \caption{An input and output example of the SQuAD dataset.}
  \label{table:example_squad}
\end{figure}

\subsubsection{Text-Pair Classification}
We also evaluate our method on cross-lingual text-pair classification tasks.
The MultiNLI dataset~\citep{williams-etal-2018-broad} involves determining the logical relationship between a premise sentence and a hypothesis sentence.
XNLI~\citep{conneau-etal-2018-xnli} is a multilingual NLI dataset for evaluation.

\begin{figure}[h]
  \begin{center}
  \begin{tabular}{p{0.95\columnwidth}}
      \toprule
\begin{alltt}
Premise: Conceptually cream skimming has two basic dimensions - product and geography.
Hypothesis: Product and geography are what make cream skimming work.
What is their logical relation? Entailment, Neutral or Contradition.

\end{alltt}
\\\midrule
\begin{alltt}
Neutral

\end{alltt}
    \\ \bottomrule
  \end{tabular}
  \end{center}
  \caption{An input and output example of the MultiNLP dataset.}
  \label{table:example_nli}
\end{figure}


\subsubsection{Mathematical Reasoning}
GSM8k~\citep{Cobbe2021TrainingVT} is an English dataset of 8.5K high-quality grade school math problems.
Each problem is annotated with a solution that shows the mathematical steps required to reach the final answer.
As the evaluation dataset, we use MGSM~\citep{DBLP:conf/iclr/ShiSF0SVCTRZ0W23}, a multilingual version of the GSM8k dataset.

The LLM is trained to generate the step-by-step solution to math problems.
The answer is extracted from the LLM output as the final digits, and the accuracy is calculated based on the exact match of the extracted answer and the ground truth.

\begin{table}[h]
  \begin{center}
  \begin{tabular}{p{0.95\columnwidth}}
    \toprule
\begin{verbatim}
Natalia sold clips to 48 of her friends in April,
and then she sold half as many clips in May.
How many clips did Natalia sell altogether in April and May?

\end{verbatim}
\\\midrule
\begin{verbatim}
Natalia sold 48/2 = clips in May.
Natalia sold 48+24 = 72 clips altogether in April and May.
#### 72

\end{verbatim}
\\ \bottomrule

  \end{tabular}
  \end{center}
  \caption{An input and output example of the GSM8k dataset.}
  \label{table:example_gsm8k}
\end{table}

\subsection{Prompt Format for LLM Translation}
\label{appendix:prompt_format}

To translate training data using a LLM (\cref{subsec:synthetic_data_generation}), we employed the following prompt template for each task.
The template simply consists of the source text and target text prepended with the language tag.
The text is surrounded by backticks and the LLM starts generating the target text the open backtick until the close backtick is found.

\begin{figure}[h]
  \begin{center}
  \begin{tabular}{p{0.95\columnwidth}}
    \toprule
    \begin{alltt}
\{% for sample in few_shot_samples %\}
en: `\{\{ sample.data_field \}\}`
\{\{ target_language \}\}: `\{\{ sample.data_field \}\}`
\{% endfor %\}
en: `\{\{ data_field \}\}`
\{\{ target_language \}\}: `
    \end{alltt}
    \\ \bottomrule
  \end{tabular}
  \end{center}
  \caption{Prompt format for LLM translation.}
  \label{table:prompt_format}
\end{figure}

The SQuAD dataset annotates answer spans in the context passages.
We translate the annotations using the mark-then-translate approach~\citep{chen-etal-2023-frustratingly}.
We mark the answer span in the context passage with the tokens ``<answer>'' and ``</answer>'', translate the marked text, and then extract the translated answer span from the translated context.
Note that in this case, the few-shot samples are also marked with the answer span.

The context passages in the SQuAD dataset are relatively long, and it is challenging for the LLM with a limited context window to fit the entire few-shot samples and the source text.
To address this issue, we split the context into sentences using the spaCy library\footnote{\url{https://spacy.io/}} and translate them separately, i.e., the few-shot samples and the source text are sentences.

\subsection{Data Filtering for Synthetic Data}
\label{appendix:data_filtering}

We remove pairs where the target length is less than one-third or more than three times the source length.
The text length is heuristically determined to account for character length differences between languages.
For example, phonogram-based text (e.g., English) has much more characters than ideograph-based text (e.g., Chinese).
We set normalization factors where English, German, Thai, and Russian characters count as 1, and Chinese characters as 3.

We also filter out incomplete translations which are typically produced by repetitive generation.
We set the maximum number of tokens for generation (\cref{table:max_number_of_tokens}) and remove the outputs not ending with the token indicating the end of the translation, in our case, the backtick character used in the prompt format.

\begin{table*}[ht]
    \centering
\begin{tabular}{lcc} \toprule
                                                                                 & Data Field & Max Number of Tokens \\ \midrule
\multirow{2}{*}{SQuAD~\citep{rajpurkar-etal-2016-squad}}   & context    & 512                  \\
                                                                                 & question   & 256                  \\ \midrule
\multirow{2}{*}{MultiNLI~\citep{williams-etal-2018-broad}} & premise    & 256                  \\
                                                                                 & hypothesis & 256                  \\ \midrule
\multirow{2}{*}{GSM8k~\citep{Cobbe2021TrainingVT}}         & question   & 512                  \\
                                                                                 & answer     & 512                  \\
\bottomrule
\end{tabular}
\caption{Maximum number of tokens set for generating translations.}
\label{table:max_number_of_tokens}
\end{table*}

\subsection{Assessing the Translation Quality}
\label{appendix:translation_quality}

To evaluate the translation quality, we use the BLEU score~\citep{papineni-etal-2002-bleu} measured by the parallel data constructed from questions in the MGSM test set.
The translation is performed in few-shot in-context learning with 8 translation samples constructed from the train set of the MGSM dataset.
The BLEU score is calculated using the \texttt{SacreBLEU} library~\citep{post-2018-call}\footnote{\url{https://github.com/mjpost/sacrebleu}}.
As the tokenizer option, we use ``13a'' for de and ru, ``flores101'' for th, and ``zh'' for zh.


\cref{tab:bleu_mgsm_appendix} shows the BLEU scores from the LLMs evaluated in this paper.
The result of \modelName{Qwen1.5-1.8B} is discussed in \cref{appendix:results_qwen}, and \modelName{gpt-3.5-turbo-0125} in \cref{appendix:gpt_results}.

\begin{table}[h]
\centering
    \setlength{\tabcolsep}{0.35em}
    \scalebox{1.0}{
    \begin{tabular}{lcccccc}
    \toprule
    \textbf{Model}      & \textbf{de}   & \textbf{ru} & \textbf{th} & \textbf{zh} \\ \midrule
    \modelName{Llama2-7B}           & 37.1          & 27.2        & 1.9         & 29.4        \\
    \modelName{Llama2-13B}          & 41.3          & 33.4        & 5.1         & 34.3        \\
    \modelName{Llama2-70B}          & 45.6          & 41.9        & 12.0        & 42.4        \\
    \modelName{Qwen1.5-1.8B}        & 21.9          & 11.3        & 1.5         & 41.3        \\
    \modelName{gpt-3.5-turbo-0125}  & 48.0          & 44.6        & 23.1        & 47.2        \\
    \bottomrule
    \end{tabular}
    }
\caption{BLEU scores from the MGSM test set.}
\label{tab:bleu_mgsm_appendix}
\end{table}

\subsection{Hyper-parameters for Fine-tuning}
\label{appendix:hyperparameters}

We provide the hyper-parameters used for fine-tuning the LLMs in \cref{table:hyperparameters}.

\begin{table*}[ht]
    \centering
\begin{tabular}{ll|ll} \toprule
\textbf{Hyper-parameter} & \textbf{Value} & \textbf{Hyper-parameter} & \textbf{Value} \\  \midrule
Batch size & 64 & Adam $\epsilon$ & 1e-8 \\
Number of steps & 1,000 & Adam $\beta_1$ & 0.9 \\
Learning rate & [5e-5, 3e-4] & Adam $\beta_2$ & 0.999 \\
LR Scheduler & Cosine & Weight decay & 0.1 \\
Warmup ratio & 0.05 & & \\
\bottomrule
\end{tabular}
\caption{Hyper-parameters used for fine-tuning the LLMs.}
\label{table:hyperparameters}
\end{table*}

\section{Results from \modelName{Qwen1.5-1.8B}}
\label{appendix:results_qwen}

To increase the robustness of the results, we also conducted experiments with \modelName{Qwen1.5-1.8B}~\footnote{\url{https://qwenlm.github.io/blog/qwen1.5/}}.
While the model is mainly trained on Chinese and English data, it is also constructed with the multilingual use cases in mind.

\begin{table*}[h!]
\centering
    \setlength{\tabcolsep}{0.40em}
    \scalebox{1.0}{
\begin{tabular}{l|cccc}
\toprule
 & \multicolumn{4}{|c}{MGSM} \\
 & de & ru & th & zh \\ \midrule
$\mathcal{D}_{\text{src}}$ & \scorewithinterval{8.0}{0.3} & \scorewithinterval{6.5}{0.4} & \scorewithinterval{2.8}{0.5} & \scorewithinterval{23.3}{0.5} \\
$+\mathcal{D}_{\text{tgt}}$ & \significantscorewithinterval{\highestscore{18.7}}{1.8} & \significantscorewithinterval{\highestscore{15.0}}{1.2} & \scorewithinterval{3.4}{1.8} & \scorewithinterval{21.4}{1.2} \\
$+\mathcal{D}_{\text{cs}}$ & \significantscorewithinterval{17.1}{1.3} & \significantscorewithinterval{14.9}{0.5} & \scorewithinterval{2.7}{0.8} & \scorewithinterval{23.2}{0.9} \\
\bottomrule
\end{tabular}
    }
\caption{Results on the MGSM dataset with \modelName{Qwen1.5-1.8B}.
    Scores are marked with $^*$ if its improvement is statistically significant ($p < 0.05$ in Welch's t-test) compared to the baseline $\mathcal{D}_{\text{src}}$.
    The significant and highest score in each column is marked in bold.
}
\label{tab:results_qwen}
\end{table*}

We observe that the results are consistent with the main experiments: Self-Translate-Train is effective when the zero-shot cross-lingual transfer performance is suboptimal and the model can generate reasonable translations.
The performance is improved by adding the target language data $+\mathcal{D}_{\text{tgt}}$.
However, when the translation quality is poor as in Thai (1.5 BLEU score in \cref{tab:bleu_mgsm_appendix}), the improvement is not observed.
Additionally, \modelName{Qwen1.5-1.8B} seems to have good cross-lingual capability between English and Chinese, as indicated by the high BLEU score (41.3 in \cref{tab:bleu_mgsm_appendix}).
With this, tuning on the source language data alone is sufficient to achieve high performance.

\section{Additional Exepriments}
\label{appendix:additional_experiments}

In this section, we discuss experiments that are outside the scope of the main topic of this paper but are somewhat relevant and may be of interest to readers.

\subsection{Does Self-Translate-Train improve the performance in the source language?}
\label{appendix:source_language}

The performance somtimes improves, given the task is challenging and the translation quality is sufficiently high.

\cref{tab:en_test_results} shows the results on the English test set with \modelName{Llama2-7B}.
The performance improves in the MGSM dataset when adding the synthetic data from de, ru, and zh.
The Thai language does not show the improvement possibly due to the low translation quality.

However, the improvement is not observed in the XQuAD and XNLI datasets.
This might be because the task performance is already high with the source language data alone, and the synthetic data does not provide additional information to improve the performance.

\begin{table*}[h!]
\centering
    \setlength{\tabcolsep}{0.40em}
    \scalebox{1.0}{
\begin{tabular}{l|cccc|cccc|cccc}
\toprule
 & \multicolumn{4}{|c|}{MGSM} & \multicolumn{4}{|c|}{XQuAD} & \multicolumn{4}{|c}{XNLI} \\
 & de & ru & th & zh & de & ru & th & zh & de & ru & th & zh \\ \midrule

$\mathcal{D}_{\text{src}}$ & \multicolumn{4}{c|}{\scorewithinterval{37.8}{0.8}} & \multicolumn{4}{|c|}{\scorewithinterval{70.2}{0.5}} & \multicolumn{4}{|c}{\scorewithinterval{88.2}{1.3}} \\
$+\mathcal{D}_{\text{tgt}}$ & \significantscorewithinterval{42.6}{1.9} & \significantscorewithinterval{41.8}{1.3} & \scorewithinterval{40.0}{1.9} & \significantscorewithinterval{\highestscore{42.7}}{1.0} & \scorewithinterval{69.1}{0.7} & \scorewithinterval{69.6}{0.5} & \scorewithinterval{70.0}{0.5} & \scorewithinterval{69.9}{0.4} & \scorewithinterval{89.0}{0.4} & \scorewithinterval{88.3}{0.7} & \scorewithinterval{88.6}{1.0} & \scorewithinterval{89.3}{0.3} \\
$+\mathcal{D}_{\text{cs}}$ & \significantscorewithinterval{\highestscore{42.7}}{0.7} & \significantscorewithinterval{\highestscore{42.9}}{0.7} & \scorewithinterval{39.8}{1.6} & \scorewithinterval{40.8}{2.0} & \scorewithinterval{69.5}{0.5} & \scorewithinterval{69.6}{0.3} & \scorewithinterval{69.8}{0.5} & \scorewithinterval{68.7}{0.5} & \scorewithinterval{88.7}{0.2} & \scorewithinterval{88.5}{1.0} & \scorewithinterval{88.0}{0.7} & \scorewithinterval{88.7}{0.6} \\
\bottomrule
\end{tabular}
    }
\caption{Results on the Englihs test set with \modelName{Llama2-7B}.
    Scores are marked with $^*$ if its improvement is statistically significant ($p < 0.05$ in Welch's t-test) compared to the baseline $\mathcal{D}_{\text{src}}$.
    The significant and highest score in each column is marked in bold.
}
\label{tab:en_test_results}
\end{table*}

\subsection{Does the synthetic data alone improve the performance in the target language?}

Yes, but adding the source language data is more effective.
\cref{tab:target_alone_results} shows the results with \modelName{Llama2-7B} on the multilingual evaluation datasets.
Tuning on the synthetic data alone ($\mathcal{D}_{\text{tgt}}$) improves the performance in the target language, but the improvement is not as significant as adding the synthetic data to the source language data ($+\mathcal{D}_{\text{tgt}}$).
In practice, we recommend using the synthetic data in combination with the original data to achieve the best performance.

\begin{table*}[h!]
\centering
    \setlength{\tabcolsep}{0.40em}
    \scalebox{0.9}{
    \begin{tabular}{l|cccc|cccc|cccc}
\toprule
 & \multicolumn{4}{|c|}{MGSM} & \multicolumn{4}{|c|}{XQuAD} & \multicolumn{4}{|c}{XNLI} \\
 & de & ru & th & zh & de & ru & th & zh & de & ru & th & zh \\ \midrule
$\mathcal{D}_{\text{src}}$ & \scorewithinterval{30.1}{0.4} & \scorewithinterval{25.0}{0.7} & \scorewithinterval{8.1}{0.7} & \scorewithinterval{21.1}{1.7} & \scorewithinterval{60.3}{0.8} & \scorewithinterval{49.3}{0.4} & \scorewithinterval{34.5}{1.0} & \scorewithinterval{66.3}{0.8} & \scorewithinterval{79.7}{0.4} & \scorewithinterval{76.9}{0.1} & \scorewithinterval{53.7}{0.9} & \scorewithinterval{74.1}{0.2} \\
\hline
$D_{\text{tgt}}$ & \significantscorewithinterval{32.1}{0.6} & \significantscorewithinterval{30.4}{1.8} & \scorewithinterval{8.7}{0.6} & \scorewithinterval{24.6}{2.5} & \significantscorewithinterval{\highestscore{62.5}}{0.4} & \significantscorewithinterval{57.4}{1.1} & \significantscorewithinterval{44.3}{0.8} & \significantscorewithinterval{75.4}{1.1} & \significantscorewithinterval{81.5}{0.6} & \scorewithinterval{77.9}{0.7} & \scorewithinterval{53.8}{1.7} & \significantscorewithinterval{77.1}{0.5} \\
$+\mathcal{D}_{\text{tgt}}$ & \significantscorewithinterval{\highestscore{36.4}}{1.3} & \significantscorewithinterval{34.0}{1.7} & \scorewithinterval{7.7}{3.4} & \significantscorewithinterval{27.1}{1.2} & \scorewithinterval{61.7}{0.9} & \significantscorewithinterval{57.8}{0.8} & \significantscorewithinterval{\highestscore{46.4}}{1.3} & \significantscorewithinterval{\highestscore{77.7}}{0.4} & \significantscorewithinterval{\highestscore{81.6}}{0.7} & \significantscorewithinterval{\highestscore{78.5}}{0.6} & \scorewithinterval{56.3}{1.5} & \significantscorewithinterval{\highestscore{78.5}}{0.3} \\
\bottomrule
\end{tabular}
    }
\caption{Results on the Englihs test set with \modelName{Llama2-7B} with the setting of tuning the synthetic data alone $\mathcal{D}_{\text{tgt}}$.}
\label{tab:target_alone_results}
\end{table*}

\subsection{What about the Translate-Test approach?}
\label{appendix:translate_test}

The translate-test approach, where non-English inputs are translated into English and fed into an English model (i.e., the model trained with $\mathcal{D}_{\text{src}}$ ), can be an alternative to the trainslate-train approach.
We perform experiments with the translate-test approach and provide additional context on the effectiveness of Self-Translate-Train.
The tasks to use are the math and NLI tasks, whose outputs are language-independent~\footnote{We cannot simply apply the translate-test approach to QA, as the output language would be different from the original language. We could remedy this by further translating the output back to the original language~\citep{Asai2018MultilingualER}, but this is out of the scope of our exploration here.}.

We first experiment with the Self-Translate-Test approach in few-shot learning settings~\citep{etxaniz-etal-2024-multilingual}.
This approach translates the test-time inputs to high-resource language (e.g., English) by few-shot translation and performs inference on the translated inputs.

We compare the Self-Translate-Test approach with a baseline without translation.
The \texttt{FewShot} baseline uses 8-shot samples in the target language from the train set for inference on the test set.
The \texttt{Self-Translate-Test (few-shot)} approach translates the test set to English and performs inference using English 8-shot samples.
The results are shown in \cref{tab:translate_test_fewshot}.

\begin{table*}[h!]
\centering
    \setlength{\tabcolsep}{0.40em}
    \begin{tabular}{l|cccc|cccc}
\toprule
 & \multicolumn{4}{|c|}{MGSM} & \multicolumn{4}{|c}{XNLI} \\
 & de & ru & th & zh & de & ru & th & zh \\ \midrule
\texttt{FewShot} & \scorewithinterval{11.2}{1.4} & \scorewithinterval{11.9}{1.5} & \scorewithinterval{3.3}{0.9} & \scorewithinterval{10.3}{0.2}  & \scorewithinterval{41.8}{0.1} & \scorewithinterval{39.0}{0.2} & \scorewithinterval{34.2}{0.0} & \scorewithinterval{38.5}{0.3} \\
\texttt{Self-Translate-Test (few-shot)} & \scorewithinterval{\highestscore{15.7}}{0.2} & \scorewithinterval{\highestscore{13.3}}{0.8} & \scorewithinterval{\highestscore{4.4}}{0.6} & \scorewithinterval{\highestscore{12.8}}{1.2} & \scorewithinterval{\highestscore{44.7}}{0.2} & \scorewithinterval{\highestscore{43.3}}{0.2} & \scorewithinterval{\highestscore{38.6}}{0.3} & \scorewithinterval{\highestscore{42.9}}{0.5} \\
\bottomrule
\end{tabular}
\caption{Results on multilingual evaluation datasets with \modelName{Llama2-7B} in the setting of few-shot learning.}
\label{tab:translate_test_fewshot}
\end{table*}

We can see that the Self-Translate-Test approach is effective both in the math and NLI tasks, which conforms to the results in \citet{etxaniz-etal-2024-multilingual}.

Next, we validate the effectiveness in fine-tuning settings.
Specifically, we feed the model fine-tuned with English task-specific data $\mathcal{D}_{\text{src}}$ with test inputs translated into English by few-shot translation of the pre-trained model.
The results are shown in \cref{tab:translate_test_finetuned}.

\begin{table*}[h!]
\centering
    \setlength{\tabcolsep}{0.40em}
    \begin{tabular}{l|cccc|cccc}
\toprule
 & \multicolumn{4}{|c|}{MGSM} & \multicolumn{4}{|c}{XNLI} \\
 & de & ru & th & zh & de & ru & th & zh \\ \midrule
$\mathcal{D}_{\text{src}}$ & \scorewithinterval{30.1}{0.4} & \scorewithinterval{25.0}{0.7} & \scorewithinterval{\highestscore{8.1}}{0.7} & \scorewithinterval{21.1}{1.7} & \scorewithinterval{79.7}{0.4} & \scorewithinterval{76.9}{0.1} & \scorewithinterval{53.7}{0.9} & \scorewithinterval{74.1}{0.2} \\
+ \texttt{Self-Translate-Test} & \scorewithinterval{\highestscore{32.2}}{1.9} & \scorewithinterval{\highestscore{32.5}}{1.9} & \scorewithinterval{7.2}{0.6} & \scorewithinterval{\highestscore{\highestscore{28.8}}}{0.3} & \scorewithinterval{\highestscore{87.3}}{1.0} & \scorewithinterval{\highestscore{86.9}}{1.0} & \scorewithinterval{\highestscore{82.4}}{1.3} & \scorewithinterval{\highestscore{86.0}}{1.3} \\
\bottomrule
\end{tabular}
\caption{Results on multilingual evaluation datasets with \modelName{Llama2-7B} in the setting of fine-tuning.}
\label{tab:translate_test_finetuned}
\end{table*}

We observe that the Self-Translate-Test approach remains effective when combined with a fine-tuned model.
This strengthen the argument that the self-translate approach is an effective way to improve the cross-lingual performance of the model under the same data budget.
However, this approach is not always practical, as it requires both the pre-trained model for few-shot translation and the fine-tuned model to be available during deployment.

\subsection{How effective is $\mathcal{D}_{\text{tgt}}$ generated by an external system?}
\label{appendix:gpt_results}

When the synthetic training data is generated by an external system, such approach can be seen as the Translate-Train approach (\cref{subsec:cross-lingual_transfer_learning}) or sequence distillation from a teacher model~\citep{kim-rush-2016-sequence}.
As an upper-bound experiment using the math task, we fine-tune \modelName{Llama2-7B} on the synthetic data generated by \modelName{gpt-3.5-turbo-0125} from the OpenAI API\footnote{\url{https://openai.com/index/openai-api/}}, which produces high-quality translations across the languages explored in this paper (\cref{tab:bleu_mgsm_appendix}).

\begin{table*}[h!]
\centering
    \setlength{\tabcolsep}{0.40em}
    \scalebox{1.0}{
\begin{tabular}{l|cccc}
\toprule
 & \multicolumn{4}{|c}{MGSM} \\
 & de & ru & th & zh \\ \midrule
$\mathcal{D}_{\text{src}}$ & \scorewithinterval{30.1}{0.4} & \scorewithinterval{25.0}{0.7} & \scorewithinterval{8.1}{0.7} & \scorewithinterval{21.1}{1.7} \\
\hline
$+\mathcal{D}_{\text{tgt}}$ from \modelName{Llama2-7B} & \significantscorewithinterval{36.4}{1.3} & \significantscorewithinterval{34.0}{1.7} & \scorewithinterval{7.7}{3.4} & \significantscorewithinterval{27.1}{1.2}  \\
$+\mathcal{D}_{\text{tgt}} + \mathcal{D}_{\text{cs}}$ from \modelName{Llama2-7B} & \significantscorewithinterval{35.9}{1.3} & \significantscorewithinterval{34.5}{2.4} & \scorewithinterval{10.4}{1.6} & \significantscorewithinterval{28.8}{1.9} \\
\hline
$+\mathcal{D}_{\text{tgt}}$ from \modelName{gpt-3.5-turbo-0125} & \significantscorewithinterval{37.4}{1.0} & \significantscorewithinterval{\highestscore{35.9}}{1.2} & \significantscorewithinterval{28.5}{1.4} & \significantscorewithinterval{33.2}{1.1} \\
$+\mathcal{D}_{\text{cs}}$ from \modelName{gpt-3.5-turbo-0125}& \significantscorewithinterval{\highestscore{38.6}}{1.5} & \significantscorewithinterval{34.4}{0.5} & \significantscorewithinterval{\highestscore{28.7}}{1.3} & \significantscorewithinterval{\highestscore{34.8}}{1.3} \\
\bottomrule
\end{tabular}
    }
\caption{Results on multilingual evaluation datasets with \modelName{Llama2-7B} tuned on the synthetic data generated by \modelName{gpt-3.5-turbo-0125}.}
\label{tab:gpt_results}
\end{table*}

The synthetic data generated by \modelName{gpt-3.5-turbo-0125} improves performance across languages over the baseline $\mathcal{D}_{\text{src}}$ and the synthetic data generated by \modelName{Llama2-7B}.
This confirms that the quality of the synthetic data is crucial for the performance improvement.

Note that the outputs from external models are often restricted in its usage\footnote{For example, Term of use of OpenAI API (January 31, 2024) restricts the usage of the outputs for training a model that competes with the API (\url{https://openai.com/policies/terms-of-use/}). Meta Llama 3 License (April 18, 2024) prohibits using the outputs to improve any other large language model (\url{https://llama.meta.com/llama3/license/}).}, while the method explored in this paper can be used with the model at hand without other resources.
Additionally, our interest in this paper is rather to explore the cross-lingual potential of the model itself and how to better utilize it.

%% file: main.bbl
\begin{thebibliography}{37}
\providecommand{\natexlab}[1]{#1}

\bibitem[{Artetxe et~al.(2023)Artetxe, Goswami, Bhosale, Fan, and
  Zettlemoyer}]{artetxe-etal-2023-revisiting}
Mikel Artetxe, Vedanuj Goswami, Shruti Bhosale, Angela Fan, and Luke
  Zettlemoyer. 2023.
\newblock \href {https://doi.org/10.18653/v1/2023.emnlp-main.399} {Revisiting
  machine translation for cross-lingual classification}.
\newblock In \emph{Proceedings of the 2023 Conference on Empirical Methods in
  Natural Language Processing}, pages 6489--6499, Singapore. Association for
  Computational Linguistics.

\bibitem[{Artetxe et~al.(2020)Artetxe, Ruder, and
  Yogatama}]{artetxe-etal-2020-cross}
Mikel Artetxe, Sebastian Ruder, and Dani Yogatama. 2020.
\newblock \href {https://doi.org/10.18653/v1/2020.acl-main.421} {On the
  cross-lingual transferability of monolingual representations}.
\newblock In \emph{Proceedings of the 58th Annual Meeting of the Association
  for Computational Linguistics}, pages 4623--4637, Online. Association for
  Computational Linguistics.

\bibitem[{Artetxe and Schwenk(2019)}]{artetxe-schwenk-2019-massively}
Mikel Artetxe and Holger Schwenk. 2019.
\newblock \href {https://doi.org/10.1162/tacl_a_00288} {Massively multilingual
  sentence embeddings for zero-shot cross-lingual transfer and beyond}.
\newblock \emph{Transactions of the Association for Computational Linguistics},
  7:597--610.

\bibitem[{Asai et~al.(2018)Asai, Eriguchi, Hashimoto, and
  Tsuruoka}]{Asai2018MultilingualER}
Akari Asai, Akiko Eriguchi, Kazuma Hashimoto, and Yoshimasa Tsuruoka. 2018.
\newblock \href {https://api.semanticscholar.org/CorpusID:52183904}
  {Multilingual extractive reading comprehension by runtime machine
  translation}.
\newblock \emph{ArXiv}, abs/1809.03275.

\bibitem[{Bai et~al.(2022)Bai, Kadavath, Kundu, Askell, Kernion, Jones, Chen,
  Goldie, Mirhoseini, McKinnon, Chen, Olsson, Olah, Hernandez, Drain, Ganguli,
  Li, Tran-Johnson, Perez, Kerr, Mueller, Ladish, Landau, Ndousse, Luko{\v
  s}iūtė, Lovitt, Sellitto, Elhage, Schiefer, Mercado, Dassarma, Lasenby,
  Larson, Ringer, Johnston, Kravec, Showk, Fort, Lanham, Telleen-Lawton,
  Conerly, Henighan, Hume, Bowman, Hatfield-Dodds, Mann, Amodei, Joseph,
  McCandlish, Brown, and Kaplan}]{Bai2022ConstitutionalAH}
Yuntao Bai, Saurav Kadavath, Sandipan Kundu, Amanda Askell, John Kernion, Andy
  Jones, Anna Chen, Anna Goldie, Azalia Mirhoseini, Cameron McKinnon, Carol
  Chen, Catherine Olsson, Christopher Olah, Danny Hernandez, Dawn Drain, Deep
  Ganguli, Dustin Li, Eli Tran-Johnson, E~Perez, Jamie Kerr, Jared Mueller,
  Jeff Ladish, J~Landau, Kamal Ndousse, Kamilė Luko{\v s}iūtė, Liane Lovitt,
  Michael Sellitto, Nelson Elhage, Nicholas Schiefer, Noem'i Mercado, Nova
  Dassarma, Robert Lasenby, Robin Larson, Sam Ringer, Scott Johnston, Shauna
  Kravec, Sheer~El Showk, Stanislav Fort, Tamera Lanham, Timothy
  Telleen-Lawton, Tom Conerly, Tom Henighan, Tristan Hume, Sam Bowman, Zac
  Hatfield-Dodds, Benjamin Mann, Dario Amodei, Nicholas Joseph, Sam McCandlish,
  Tom~B. Brown, and Jared Kaplan. 2022.
\newblock \href {https://api.semanticscholar.org/CorpusID:254823489}
  {Constitutional ai: Harmlessness from ai feedback}.
\newblock \emph{ArXiv}, abs/2212.08073.

\bibitem[{Briakou et~al.(2023)Briakou, Cherry, and
  Foster}]{briakou-etal-2023-searching}
Eleftheria Briakou, Colin Cherry, and George Foster. 2023.
\newblock \href {https://doi.org/10.18653/v1/2023.acl-long.524} {Searching for
  needles in a haystack: On the role of incidental bilingualism in {P}a{LM}{'}s
  translation capability}.
\newblock In \emph{Proceedings of the 61st Annual Meeting of the Association
  for Computational Linguistics (Volume 1: Long Papers)}, pages 9432--9452,
  Toronto, Canada. Association for Computational Linguistics.

\bibitem[{Brown et~al.(2020)Brown, Mann, Ryder, Subbiah, Kaplan, Dhariwal,
  Neelakantan, Shyam, Sastry, Askell, Agarwal, {Herbert-Voss}, Krueger,
  Henighan, Child, Ramesh, Ziegler, Wu, Winter, Hesse, Chen, Sigler, Litwin,
  Gray, Chess, Clark, Berner, McCandlish, Radford, Sutskever, and
  Amodei}]{brownLanguageModelsAre2020}
Tom Brown, Benjamin Mann, Nick Ryder, Melanie Subbiah, Jared~D Kaplan, Prafulla
  Dhariwal, Arvind Neelakantan, Pranav Shyam, Girish Sastry, Amanda Askell,
  Sandhini Agarwal, Ariel {Herbert-Voss}, Gretchen Krueger, Tom Henighan, Rewon
  Child, Aditya Ramesh, Daniel Ziegler, Jeffrey Wu, Clemens Winter, Chris
  Hesse, Mark Chen, Eric Sigler, Mateusz Litwin, Scott Gray, Benjamin Chess,
  Jack Clark, Christopher Berner, Sam McCandlish, Alec Radford, Ilya Sutskever,
  and Dario Amodei. 2020.
\newblock Language {{Models}} are {{Few-Shot Learners}}.
\newblock In \emph{Advances in {{Neural Information Processing Systems}}},
  volume~33, pages 1877--1901. {Curran Associates, Inc.}

\bibitem[{Chen et~al.(2021)Chen, Ma, Chen, Dong, Zhang, Pan, Wang, and
  Wei}]{chen-etal-2021-zero}
Guanhua Chen, Shuming Ma, Yun Chen, Li~Dong, Dongdong Zhang, Jia Pan, Wenping
  Wang, and Furu Wei. 2021.
\newblock \href {https://doi.org/10.18653/v1/2021.emnlp-main.2} {Zero-shot
  cross-lingual transfer of neural machine translation with multilingual
  pretrained encoders}.
\newblock In \emph{Proceedings of the 2021 Conference on Empirical Methods in
  Natural Language Processing}, pages 15--26, Online and Punta Cana, Dominican
  Republic. Association for Computational Linguistics.

\bibitem[{Chen et~al.(2023)Chen, Jiang, Ritter, and
  Xu}]{chen-etal-2023-frustratingly}
Yang Chen, Chao Jiang, Alan Ritter, and Wei Xu. 2023.
\newblock \href {https://doi.org/10.18653/v1/2023.findings-acl.357}
  {Frustratingly easy label projection for cross-lingual transfer}.
\newblock In \emph{Findings of the Association for Computational Linguistics:
  ACL 2023}, pages 5775--5796, Toronto, Canada. Association for Computational
  Linguistics.

\bibitem[{Cobbe et~al.(2021)Cobbe, Kosaraju, Bavarian, Chen, Jun, Kaiser,
  Plappert, Tworek, Hilton, Nakano, Hesse, and Schulman}]{Cobbe2021TrainingVT}
Karl Cobbe, Vineet Kosaraju, Mohammad Bavarian, Mark Chen, Heewoo Jun, Lukasz
  Kaiser, Matthias Plappert, Jerry Tworek, Jacob Hilton, Reiichiro Nakano,
  Christopher Hesse, and John Schulman. 2021.
\newblock \href {https://api.semanticscholar.org/CorpusID:239998651} {Training
  verifiers to solve math word problems}.
\newblock \emph{ArXiv}, abs/2110.14168.

\bibitem[{Conneau et~al.(2020)Conneau, Khandelwal, Goyal, Chaudhary, Wenzek,
  Guzm{\'a}n, Grave, Ott, Zettlemoyer, and
  Stoyanov}]{conneau-etal-2020-unsupervised}
Alexis Conneau, Kartikay Khandelwal, Naman Goyal, Vishrav Chaudhary, Guillaume
  Wenzek, Francisco Guzm{\'a}n, Edouard Grave, Myle Ott, Luke Zettlemoyer, and
  Veselin Stoyanov. 2020.
\newblock \href {https://doi.org/10.18653/v1/2020.acl-main.747} {Unsupervised
  {{Cross-lingual Representation Learning}} at {{Scale}}}.
\newblock In \emph{Proceedings of the 58th {{Annual Meeting}} of the
  {{Association}} for {{Computational Linguistics}}}, pages 8440--8451.
  {Association for Computational Linguistics}.

\bibitem[{Conneau et~al.(2018)Conneau, Rinott, Lample, Williams, Bowman,
  Schwenk, and Stoyanov}]{conneau-etal-2018-xnli}
Alexis Conneau, Ruty Rinott, Guillaume Lample, Adina Williams, Samuel Bowman,
  Holger Schwenk, and Veselin Stoyanov. 2018.
\newblock \href {https://doi.org/10.18653/v1/D18-1269} {{XNLI}: Evaluating
  cross-lingual sentence representations}.
\newblock In \emph{Proceedings of the 2018 Conference on Empirical Methods in
  Natural Language Processing}, pages 2475--2485, Brussels, Belgium.
  Association for Computational Linguistics.

\bibitem[{Ebing and Glava{\v{s}}(2024)}]{ebing-glavas-2024-translate}
Benedikt Ebing and Goran Glava{\v{s}}. 2024.
\newblock \href {https://doi.org/10.18653/v1/2024.naacl-long.298} {To translate
  or not to translate: A systematic investigation of translation-based
  cross-lingual transfer to low-resource languages}.
\newblock In \emph{Proceedings of the 2024 Conference of the North American
  Chapter of the Association for Computational Linguistics: Human Language
  Technologies (Volume 1: Long Papers)}, pages 5325--5344, Mexico City, Mexico.
  Association for Computational Linguistics.

\bibitem[{Ebrahimi and von~der Wense(2024)}]{ebrahimi-wense-2024-zero}
Abteen Ebrahimi and Katharina von~der Wense. 2024.
\newblock \href {https://doi.org/10.18653/v1/2024.naacl-short.37} {Zero-shot
  vs. translation-based cross-lingual transfer: The case of lexical gaps}.
\newblock In \emph{Proceedings of the 2024 Conference of the North American
  Chapter of the Association for Computational Linguistics: Human Language
  Technologies (Volume 2: Short Papers)}, pages 443--458, Mexico City, Mexico.
  Association for Computational Linguistics.

\bibitem[{Etxaniz et~al.(2024)Etxaniz, Azkune, Soroa, Lacalle, and
  Artetxe}]{etxaniz-etal-2024-multilingual}
Julen Etxaniz, Gorka Azkune, Aitor Soroa, Oier Lacalle, and Mikel Artetxe.
  2024.
\newblock \href {https://doi.org/10.18653/v1/2024.naacl-short.46} {Do
  multilingual language models think better in {E}nglish?}
\newblock In \emph{Proceedings of the 2024 Conference of the North American
  Chapter of the Association for Computational Linguistics: Human Language
  Technologies (Volume 2: Short Papers)}, pages 550--564, Mexico City, Mexico.
  Association for Computational Linguistics.

\bibitem[{Holtzman et~al.(2020)Holtzman, Buys, Du, Forbes, and
  Choi}]{DBLP:conf/iclr/HoltzmanBDFC20}
Ari Holtzman, Jan Buys, Li~Du, Maxwell Forbes, and Yejin Choi. 2020.
\newblock \href {https://openreview.net/forum?id=rygGQyrFvH} {The curious case
  of neural text degeneration}.
\newblock In \emph{8th International Conference on Learning Representations,
  {ICLR} 2020, Addis Ababa, Ethiopia, April 26-30, 2020}. OpenReview.net.

\bibitem[{Hu et~al.(2022)Hu, Shen, Wallis, Allen{-}Zhu, Li, Wang, Wang, and
  Chen}]{DBLP:conf/iclr/HuSWALWWC22}
Edward~J. Hu, Yelong Shen, Phillip Wallis, Zeyuan Allen{-}Zhu, Yuanzhi Li,
  Shean Wang, Lu~Wang, and Weizhu Chen. 2022.
\newblock \href {https://openreview.net/forum?id=nZeVKeeFYf9} {Lora: Low-rank
  adaptation of large language models}.
\newblock In \emph{The Tenth International Conference on Learning
  Representations, {ICLR} 2022, Virtual Event, April 25-29, 2022}.
  OpenReview.net.

\bibitem[{Hu et~al.(2020)Hu, Ruder, Siddhant, Neubig, Firat, and
  Johnson}]{pmlr-v119-hu20b}
Junjie Hu, Sebastian Ruder, Aditya Siddhant, Graham Neubig, Orhan Firat, and
  Melvin Johnson. 2020.
\newblock \href {https://proceedings.mlr.press/v119/hu20b.html} {{XTREME}: A
  massively multilingual multi-task benchmark for evaluating cross-lingual
  generalisation}.
\newblock In \emph{Proceedings of the 37th International Conference on Machine
  Learning}, volume 119 of \emph{Proceedings of Machine Learning Research},
  pages 4411--4421. PMLR.

\bibitem[{Huang et~al.(2023)Huang, Gu, Hou, Wu, Wang, Yu, and
  Han}]{huang-etal-2023-large}
Jiaxin Huang, Shixiang Gu, Le~Hou, Yuexin Wu, Xuezhi Wang, Hongkun Yu, and
  Jiawei Han. 2023.
\newblock \href {https://doi.org/10.18653/v1/2023.emnlp-main.67} {Large
  language models can self-improve}.
\newblock In \emph{Proceedings of the 2023 Conference on Empirical Methods in
  Natural Language Processing}, pages 1051--1068, Singapore. Association for
  Computational Linguistics.

\bibitem[{Kim and Rush(2016)}]{kim-rush-2016-sequence}
Yoon Kim and Alexander~M. Rush. 2016.
\newblock \href {https://doi.org/10.18653/v1/D16-1139} {Sequence-level
  knowledge distillation}.
\newblock In \emph{Proceedings of the 2016 Conference on Empirical Methods in
  Natural Language Processing}, pages 1317--1327, Austin, Texas. Association
  for Computational Linguistics.

\bibitem[{Lee et~al.(2024)Lee, Dai, Ren, Chen, Cer, Cole, Hui, Boratko,
  Kapadia, Ding, Luan, Duddu, Abrego, Shi, Gupta, Kusupati, Jain, Jonnalagadda,
  Chang, and Naim}]{Lee2024GeckoVT}
Jinhyuk Lee, Zhuyun Dai, Xiaoqi Ren, Blair Chen, Daniel Cer, Jeremy~R. Cole,
  Kai Hui, Michael Boratko, Rajvi Kapadia, Wen Ding, Yi~Luan, Sai Meher~Karthik
  Duddu, Gustavo~Hern{\'a}ndez Abrego, Weiqiang Shi, Nithi Gupta, Aditya
  Kusupati, Prateek Jain, Siddhartha~R. Jonnalagadda, Ming-Wei Chang, and
  Iftekhar Naim. 2024.
\newblock \href {https://api.semanticscholar.org/CorpusID:268793455} {Gecko:
  Versatile text embeddings distilled from large language models}.
\newblock \emph{ArXiv}, abs/2403.20327.

\bibitem[{Li et~al.(2023{\natexlab{a}})Li, Yu, Zhou, Schick, Zettlemoyer, Levy,
  Weston, and Lewis}]{DBLP:journals/corr/abs-2308-06259}
Xian Li, Ping Yu, Chunting Zhou, Timo Schick, Luke Zettlemoyer, Omer Levy,
  Jason Weston, and Mike Lewis. 2023{\natexlab{a}}.
\newblock \href {https://doi.org/10.48550/ARXIV.2308.06259} {Self-alignment
  with instruction backtranslation}.
\newblock \emph{CoRR}, abs/2308.06259.

\bibitem[{Li et~al.(2023{\natexlab{b}})Li, Zhu, Lu, and
  Yin}]{li-etal-2023-synthetic}
Zhuoyan Li, Hangxiao Zhu, Zhuoran Lu, and Ming Yin. 2023{\natexlab{b}}.
\newblock \href {https://doi.org/10.18653/v1/2023.emnlp-main.647} {Synthetic
  data generation with large language models for text classification: Potential
  and limitations}.
\newblock In \emph{Proceedings of the 2023 Conference on Empirical Methods in
  Natural Language Processing}, pages 10443--10461, Singapore. Association for
  Computational Linguistics.

\bibitem[{Loshchilov and Hutter(2019)}]{DBLP:conf/iclr/LoshchilovH19}
Ilya Loshchilov and Frank Hutter. 2019.
\newblock \href {https://openreview.net/forum?id=Bkg6RiCqY7} {Decoupled weight
  decay regularization}.
\newblock In \emph{7th International Conference on Learning Representations,
  {ICLR} 2019, New Orleans, LA, USA, May 6-9, 2019}. OpenReview.net.

\bibitem[{Mulcaire et~al.(2019)Mulcaire, Kasai, and
  Smith}]{mulcaire-etal-2019-polyglot}
Phoebe Mulcaire, Jungo Kasai, and Noah~A. Smith. 2019.
\newblock \href {https://doi.org/10.18653/v1/N19-1392} {Polyglot contextual
  representations improve crosslingual transfer}.
\newblock In \emph{Proceedings of the 2019 Conference of the North {A}merican
  Chapter of the Association for Computational Linguistics: Human Language
  Technologies, Volume 1 (Long and Short Papers)}, pages 3912--3918,
  Minneapolis, Minnesota. Association for Computational Linguistics.

\bibitem[{Papineni et~al.(2002)Papineni, Roukos, Ward, and
  Zhu}]{papineni-etal-2002-bleu}
Kishore Papineni, Salim Roukos, Todd Ward, and Wei-Jing Zhu. 2002.
\newblock \href {https://doi.org/10.3115/1073083.1073135} {{B}leu: a method for
  automatic evaluation of machine translation}.
\newblock In \emph{Proceedings of the 40th Annual Meeting of the Association
  for Computational Linguistics}, pages 311--318, Philadelphia, Pennsylvania,
  USA. Association for Computational Linguistics.

\bibitem[{Pikuliak et~al.(2021)Pikuliak, Simko, and
  Bielikov{\'a}}]{Pikuliak2021CrosslingualLF}
Mat{\'u}{\u s} Pikuliak, Mari{\'a}n Simko, and M{\'a}ria Bielikov{\'a}. 2021.
\newblock \href {https://api.semanticscholar.org/CorpusID:224879602}
  {Cross-lingual learning for text processing: A survey}.
\newblock \emph{Expert Systems with Applications}, 165:113765.

\bibitem[{Pires et~al.(2019)Pires, Schlinger, and
  Garrette}]{pires-etal-2019-multilingual}
Telmo Pires, Eva Schlinger, and Dan Garrette. 2019.
\newblock \href {https://doi.org/10.18653/v1/P19-1493} {How multilingual is
  multilingual {BERT}?}
\newblock In \emph{Proceedings of the 57th Annual Meeting of the Association
  for Computational Linguistics}, pages 4996--5001, Florence, Italy.
  Association for Computational Linguistics.

\bibitem[{Post(2018)}]{post-2018-call}
Matt Post. 2018.
\newblock \href {https://doi.org/10.18653/v1/W18-6319} {A call for clarity in
  reporting {BLEU} scores}.
\newblock In \emph{Proceedings of the Third Conference on Machine Translation:
  Research Papers}, pages 186--191, Brussels, Belgium. Association for
  Computational Linguistics.

\bibitem[{Rajpurkar et~al.(2016)Rajpurkar, Zhang, Lopyrev, and
  Liang}]{rajpurkar-etal-2016-squad}
Pranav Rajpurkar, Jian Zhang, Konstantin Lopyrev, and Percy Liang. 2016.
\newblock \href {https://doi.org/10.18653/v1/D16-1264} {{SQ}u{AD}: 100,000+
  questions for machine comprehension of text}.
\newblock In \emph{Proceedings of the 2016 Conference on Empirical Methods in
  Natural Language Processing}, pages 2383--2392, Austin, Texas. Association
  for Computational Linguistics.

\bibitem[{Scao et~al.(2022)Scao, Fan, Akiki, Pavlick, Ili'c, Hesslow,
  Castagn'e, Luccioni, Yvon, Gall{\'e}, Tow, Rush, Biderman, Webson,
  Ammanamanchi, Wang, Sagot, Muennighoff, del Moral, Ruwase, Bawden, Bekman,
  McMillan-Major, Beltagy, Nguyen, Saulnier, Tan, Suarez, Sanh, Laurenccon,
  Jernite, Launay, Mitchell, Raffel, Gokaslan, Simhi, Etxabe, Aji, Alfassy,
  Rogers, Nitzav, Xu, Mou, Emezue, Klamm, Leong, van Strien, Adelani, Radev,
  Ponferrada, Levkovizh, Kim, Natan, Toni, Dupont, Kruszewski, Pistilli,
  ElSahar, Benyamina, Tran, Yu, Abdulmumin, Johnson, Gonzalez-Dios, de~la Rosa,
  Chim, Dodge, Zhu, Chang, Frohberg, Tobing, Bhattacharjee, Almubarak, Chen,
  Lo, von Werra, Weber, Phan, Allal, Tanguy, Dey, Mu{\~n}oz, Masoud, Grandury,
  vSavsko, Huang, Coavoux, Singh, Jiang, Vu, Jauhar, Ghaleb, Subramani,
  Kassner, Khamis, Nguyen, Espejel, de~Gibert, Villegas, Henderson, Colombo,
  Amuok, Lhoest, Harliman, Bommasani, L'opez, Ribeiro, Osei, Pyysalo, Nagel,
  Bose, Muhammad, Sharma, Longpre, Nikpoor, Silberberg, Pai, Zink, Torrent,
  Schick, Thrush, Danchev, Nikoulina, Laippala, Lepercq, Prabhu, Alyafeai,
  Talat, Raja, Heinzerling, Si, Salesky, Mielke, Lee, Sharma, Santilli,
  Chaffin, Stiegler, Datta, Szczechla, Chhablani, Wang, Pandey, Strobelt,
  Fries, Rozen, Gao, Sutawika, Bari, Al-Shaibani, Manica, Nayak, Teehan,
  Albanie, Shen, Ben-David, Bach, Kim, Bers, F{\'e}vry, Neeraj, Thakker,
  Raunak, Tang, Yong, Sun, Brody, Uri, Tojarieh, Roberts, Chung, Tae, Phang,
  Press, Li, Narayanan, Bourfoune, Casper, Rasley, Ryabinin, Mishra, Zhang,
  Shoeybi, Peyrounette, Patry, Tazi, Sanseviero, von Platen, Cornette,
  Lavall'ee, Lacroix, Rajbhandari, Gandhi, Smith, Requena, Patil, Dettmers,
  Baruwa, Singh, Cheveleva, Ligozat, Subramonian, N'ev'eol, Lovering, Garrette,
  Tunuguntla, Reiter, Taktasheva, Voloshina, Bogdanov, Winata, Schoelkopf,
  Kalo, Novikova, Forde, Tang, Kasai, Kawamura, Hazan, Carpuat, Clinciu, Kim,
  Cheng, Serikov, Antverg, van~der Wal, Zhang, Zhang, Gehrmann, Mirkin, Pais,
  Shavrina, Scialom, Yun, Limisiewicz, Rieser, Protasov, Mikhailov,
  Pruksachatkun, Belinkov, Bamberger, Kasner, Kasner, Pestana, Feizpour, Khan,
  Faranak, Santos, Hevia, Unldreaj, Aghagol, Abdollahi, Tammour, HajiHosseini,
  Behroozi, Ajibade, Saxena, Ferrandis, Contractor, Lansky, David, Kiela,
  Nguyen, Tan, Baylor, Ozoani, Mirza, Ononiwu, Rezanejad, Jones, Bhattacharya,
  Solaiman, Sedenko, Nejadgholi, Passmore, Seltzer, Sanz, Fort, Dutra,
  Samagaio, Elbadri, Mieskes, Gerchick, Akinlolu, McKenna, Qiu, Ghauri,
  Burynok, Abrar, Rajani, Elkott, Fahmy, Samuel, An, Kromann, Hao, Alizadeh,
  Shubber, Wang, Roy, Viguier, Le, Oyebade, Le, Yang, Nguyen, Kashyap,
  Palasciano, Callahan, Shukla, Miranda-Escalada, Singh, Beilharz, Wang,
  de~Brito, Zhou, Jain, Xu, Fourrier, Perin'an, Molano, Yu, Manjavacas, Barth,
  Fuhrimann, Altay, Bayrak, Burns, Vrabec, Bello, Dash, Kang, Giorgi, Golde,
  Posada, Sivaraman, Bulchandani, Liu, Shinzato, de~Bykhovetz, Takeuchi,
  P{\`a}mies, Castillo, Nezhurina, Sanger, Samwald, Cullan, Weinberg, Wolf,
  Mihaljcic, Liu, Freidank, Kang, Seelam, Dahlberg, Broad, Muellner, Fung,
  Haller, Chandrasekhar, Eisenberg, Martin, Canalli, Su, Su, Cahyawijaya,
  Garda, Deshmukh, Mishra, Kiblawi, Ott, Sang-aroonsiri, Kumar, Schweter,
  Bharati, Laud, Gigant, Kainuma, Kusa, Labrak, Bajaj, Venkatraman, Xu, Xu, Xu,
  Tan, Xie, Ye, Bras, Belkada, and Wolf}]{Scao2022BLOOMA1}
Teven~Le Scao, Angela Fan, Christopher Akiki, Ellie Pavlick, Suzana Ili'c,
  Daniel Hesslow, Roman Castagn'e, Alexandra~Sasha Luccioni, François Yvon,
  Matthias Gall{\'e}, Jonathan Tow, Alexander~M. Rush, Stella Biderman, Albert
  Webson, Pawan~Sasanka Ammanamanchi, Thomas Wang, Beno{\^i}t Sagot, Niklas
  Muennighoff, Albert~Villanova del Moral, Olatunji Ruwase, Rachel Bawden, Stas
  Bekman, Angelina McMillan-Major, Iz~Beltagy, Huu Nguyen, Lucile Saulnier,
  Samson Tan, Pedro~Ortiz Suarez, Victor Sanh, Hugo Laurenccon, Yacine Jernite,
  Julien Launay, Margaret Mitchell, Colin Raffel, Aaron Gokaslan, Adi Simhi,
  Aitor~Soroa Etxabe, Alham~Fikri Aji, Amit Alfassy, Anna Rogers,
  Ariel~Kreisberg Nitzav, Canwen Xu, Chenghao Mou, Chris~C. Emezue, Christopher
  Klamm, Colin Leong, Daniel~Alexander van Strien, David~Ifeoluwa Adelani,
  Dragomir~R. Radev, Eduardo~Gonz'alez Ponferrada, Efrat Levkovizh, Ethan Kim,
  Eyal Natan, Francesco~De Toni, G{\'e}rard Dupont, Germ{\'a}n Kruszewski,
  Giada Pistilli, Hady ElSahar, Hamza Benyamina, Hieu~Trung Tran, Ian Yu, Idris
  Abdulmumin, Isaac Johnson, Itziar Gonzalez-Dios, Javier de~la Rosa, Jenny
  Chim, Jesse Dodge, Jian Zhu, Jonathan Chang, Jorg Frohberg, Josephine~L.
  Tobing, Joydeep Bhattacharjee, Khalid Almubarak, Kimbo Chen, Kyle Lo, Leandro
  von Werra, Leon Weber, Long Phan, Loubna~Ben Allal, Ludovic Tanguy, Manan
  Dey, Manuel~Romero Mu{\~n}oz, Maraim Masoud, Mar{\'i}a Grandury, Mario
  vSavsko, Max Huang, Maximin Coavoux, Mayank Singh, Mike Tian-Jian Jiang,
  Minh~Chien Vu, Mohammad~A. Jauhar, Mustafa Ghaleb, Nishant Subramani, Nora
  Kassner, Nurulaqilla Khamis, Olivier Nguyen, Omar Espejel, Ona de~Gibert,
  Paulo Villegas, Peter Henderson, Pierre Colombo, Priscilla Amuok, Quentin
  Lhoest, Rheza Harliman, Rishi Bommasani, Roberto L'opez, Rui Ribeiro, Salomey
  Osei, Sampo Pyysalo, Sebastian Nagel, Shamik Bose, Shamsuddeen~Hassan
  Muhammad, Shanya Sharma, S.~Longpre, Somaieh Nikpoor, S.~Silberberg, Suhas
  Pai, Sydney Zink, Tiago~Timponi Torrent, Timo Schick, Tristan Thrush,
  Valentin Danchev, Vassilina Nikoulina, Veronika Laippala, Violette Lepercq,
  Vrinda Prabhu, Zaid Alyafeai, Zeerak Talat, Arun Raja, Benjamin Heinzerling,
  Chenglei Si, Elizabeth Salesky, Sabrina~J. Mielke, Wilson~Y. Lee, Abheesht
  Sharma, Andrea Santilli, Antoine Chaffin, Arnaud Stiegler, Debajyoti Datta,
  Eliza Szczechla, Gunjan Chhablani, Han Wang, Harshit Pandey, Hendrik
  Strobelt, Jason~Alan Fries, Jos Rozen, Leo Gao, Lintang Sutawika, M~Saiful
  Bari, Maged~S. Al-Shaibani, Matteo Manica, Nihal~V. Nayak, Ryan Teehan,
  Samuel Albanie, Sheng Shen, Srulik Ben-David, Stephen~H. Bach, Taewoon Kim,
  Tali Bers, Thibault F{\'e}vry, Trishala Neeraj, Urmish Thakker, Vikas Raunak,
  Xiang Tang, Zheng-Xin Yong, Zhiqing Sun, Shaked Brody, Y~Uri, Hadar Tojarieh,
  Adam Roberts, Hyung~Won Chung, Jaesung Tae, Jason Phang, Ofir Press, Conglong
  Li, Deepak Narayanan, Hatim Bourfoune, Jared Casper, Jeff Rasley, Max
  Ryabinin, Mayank Mishra, Minjia Zhang, Mohammad Shoeybi, Myriam Peyrounette,
  Nicolas Patry, Nouamane Tazi, Omar Sanseviero, Patrick von Platen, Pierre
  Cornette, Pierre~Franccois Lavall'ee, R{\'e}mi Lacroix, Samyam Rajbhandari,
  Sanchit Gandhi, Shaden Smith, St{\'e}phane Requena, Suraj Patil, Tim
  Dettmers, Ahmed Baruwa, Amanpreet Singh, Anastasia Cheveleva, Anne-Laure
  Ligozat, Arjun Subramonian, Aur'elie N'ev'eol, Charles Lovering, Daniel~H
  Garrette, Deepak~R. Tunuguntla, Ehud Reiter, Ekaterina Taktasheva, Ekaterina
  Voloshina, Eli Bogdanov, Genta~Indra Winata, Hailey Schoelkopf, Jan-Christoph
  Kalo, Jekaterina Novikova, Jessica~Zosa Forde, Xiangru Tang, Jungo Kasai, Ken
  Kawamura, Liam Hazan, Marine Carpuat, Miruna Clinciu, Najoung Kim, Newton
  Cheng, Oleg Serikov, Omer Antverg, Oskar van~der Wal, Rui Zhang, Ruochen
  Zhang, Sebastian Gehrmann, Shachar Mirkin, S.~Osher Pais, Tatiana Shavrina,
  Thomas Scialom, Tian Yun, Tomasz Limisiewicz, Verena Rieser, Vitaly Protasov,
  Vladislav Mikhailov, Yada Pruksachatkun, Yonatan Belinkov, Zachary Bamberger,
  Zdenvek Kasner, Zdeněk Kasner, Amanda Pestana, Amir Feizpour, Ammar Khan,
  Amy Faranak, Ananda Santa~Rosa Santos, Anthony Hevia, Antigona Unldreaj,
  Arash Aghagol, Arezoo Abdollahi, Aycha Tammour, Azadeh HajiHosseini, Bahareh
  Behroozi, Benjamin~Ayoade Ajibade, Bharat~Kumar Saxena, Carlos~Mu{\~n}oz
  Ferrandis, Danish Contractor, David~M. Lansky, Davis David, Douwe Kiela,
  Duong~Anh Nguyen, Edward Tan, Emi Baylor, Ezinwanne Ozoani, Fatim~Tahirah
  Mirza, Frankline Ononiwu, Habib Rezanejad, H.A. Jones, Indrani Bhattacharya,
  Irene Solaiman, Irina Sedenko, Isar Nejadgholi, Jan Passmore, Joshua Seltzer,
  Julio~Bonis Sanz, Karen Fort, L{\'i}via Dutra, Mairon Samagaio, Maraim
  Elbadri, Margot Mieskes, Marissa Gerchick, Martha Akinlolu, Michael McKenna,
  Mike Qiu, Muhammed Ghauri, Mykola Burynok, Nafis Abrar, Nazneen Rajani, Nour
  Elkott, Nourhan Fahmy, Olanrewaju Samuel, Ran An, R.~P. Kromann, Ryan Hao,
  Samira Alizadeh, Sarmad Shubber, Silas~L. Wang, Sourav Roy, Sylvain Viguier,
  Thanh-Cong Le, Tobi Oyebade, Trieu Nguyen~Hai Le, Yoyo Yang, Zach Nguyen,
  Abhinav~Ramesh Kashyap, Alfredo Palasciano, Alison Callahan, Anima Shukla,
  Antonio Miranda-Escalada, Ayush~Kumar Singh, Benjamin Beilharz, Bo~Wang, Caio
  Matheus~Fonseca de~Brito, Chenxi Zhou, Chirag Jain, Chuxin Xu, Cl{\'e}mentine
  Fourrier, Daniel~Le'on Perin'an, Daniel Molano, Dian Yu, Enrique Manjavacas,
  Fabio Barth, Florian Fuhrimann, Gabriel Altay, Giyaseddin Bayrak, Gully
  Burns, Helena~U. Vrabec, Iman~I.B. Bello, Isha Dash, Ji~Soo Kang, John
  Giorgi, Jonas Golde, Jose~David Posada, Karthi Sivaraman, Lokesh Bulchandani,
  Lu~Liu, Luisa Shinzato, Madeleine~Hahn de~Bykhovetz, Maiko Takeuchi, Marc
  P{\`a}mies, Mar{\'i}a~Andrea Castillo, Marianna Nezhurina, Mario Sanger,
  Matthias Samwald, Michael Cullan, Michael Weinberg, M~Wolf, Mina Mihaljcic,
  Minna Liu, Moritz Freidank, Myungsun Kang, Natasha Seelam, Nathan Dahlberg,
  Nicholas~Michio Broad, Nikolaus Muellner, Pascale Fung, Patricia Haller,
  R.~Chandrasekhar, Renata Eisenberg, Robert Martin, Rodrigo Canalli, Rosaline
  Su, Ruisi Su, Samuel Cahyawijaya, Samuele Garda, Shlok~S Deshmukh, Shubhanshu
  Mishra, Sid Kiblawi, Simon Ott, Sinee Sang-aroonsiri, Srishti Kumar, Stefan
  Schweter, Sushil~Pratap Bharati, Tanmay Laud, Th{\'e}o Gigant, Tomoya
  Kainuma, Wojciech Kusa, Yanis Labrak, Yashasvi Bajaj, Y.~Venkatraman, Yifan
  Xu, Ying Xu, Yu~Xu, Zhee~Xao Tan, Zhongli Xie, Zifan Ye, Mathilde Bras,
  Younes Belkada, and Thomas Wolf. 2022.
\newblock \href {https://api.semanticscholar.org/CorpusID:253420279} {Bloom: A
  176b-parameter open-access multilingual language model}.
\newblock \emph{ArXiv}, abs/2211.05100.

\bibitem[{Shi et~al.(2023)Shi, Suzgun, Freitag, Wang, Srivats, Vosoughi, Chung,
  Tay, Ruder, Zhou, Das, and Wei}]{DBLP:conf/iclr/ShiSF0SVCTRZ0W23}
Freda Shi, Mirac Suzgun, Markus Freitag, Xuezhi Wang, Suraj Srivats, Soroush
  Vosoughi, Hyung~Won Chung, Yi~Tay, Sebastian Ruder, Denny Zhou, Dipanjan Das,
  and Jason Wei. 2023.
\newblock \href {https://openreview.net/pdf?id=fR3wGCk-IXp} {Language models
  are multilingual chain-of-thought reasoners}.
\newblock In \emph{The Eleventh International Conference on Learning
  Representations, {ICLR} 2023, Kigali, Rwanda, May 1-5, 2023}. OpenReview.net.

\bibitem[{Sun et~al.(2023)Sun, Shen, Zhou, Zhang, Chen, Cox, Yang, and
  Gan}]{DBLP:conf/nips/SunSZZCCYG23}
Zhiqing Sun, Yikang Shen, Qinhong Zhou, Hongxin Zhang, Zhenfang Chen, David~D.
  Cox, Yiming Yang, and Chuang Gan. 2023.
\newblock \href
  {http://papers.nips.cc/paper\_files/paper/2023/hash/0764db1151b936aca59249e2c1386101-Abstract-Conference.html}
  {Principle-driven self-alignment of language models from scratch with minimal
  human supervision}.
\newblock In \emph{Advances in Neural Information Processing Systems 36: Annual
  Conference on Neural Information Processing Systems 2023, NeurIPS 2023, New
  Orleans, LA, USA, December 10 - 16, 2023}.

\bibitem[{Touvron et~al.(2023)Touvron, Martin, Stone, Albert, Almahairi,
  Babaei, Bashlykov, Batra, Bhargava, Bhosale, Bikel, Blecher, Ferrer, Chen,
  Cucurull, Esiobu, Fernandes, Fu, Fu, Fuller, Gao, Goswami, Goyal, Hartshorn,
  Hosseini, Hou, Inan, Kardas, Kerkez, Khabsa, Kloumann, Korenev, Koura,
  Lachaux, Lavril, Lee, Liskovich, Lu, Mao, Martinet, Mihaylov, Mishra,
  Molybog, Nie, Poulton, Reizenstein, Rungta, Saladi, Schelten, Silva, Smith,
  Subramanian, Tan, Tang, Taylor, Williams, Kuan, Xu, Yan, Zarov, Zhang, Fan,
  Kambadur, Narang, Rodriguez, Stojnic, Edunov, and
  Scialom}]{Touvron2023Llama2O}
Hugo Touvron, Louis Martin, Kevin~R. Stone, Peter Albert, Amjad Almahairi,
  Yasmine Babaei, Nikolay Bashlykov, Soumya Batra, Prajjwal Bhargava, Shruti
  Bhosale, Daniel~M. Bikel, Lukas Blecher, Cristian~Cant{\'o}n Ferrer, Moya
  Chen, Guillem Cucurull, David Esiobu, Jude Fernandes, Jeremy Fu, Wenyin Fu,
  Brian Fuller, Cynthia Gao, Vedanuj Goswami, Naman Goyal, Anthony~S.
  Hartshorn, Saghar Hosseini, Rui Hou, Hakan Inan, Marcin Kardas, Viktor
  Kerkez, Madian Khabsa, Isabel~M. Kloumann, A.~V. Korenev, Punit~Singh Koura,
  Marie-Anne Lachaux, Thibaut Lavril, Jenya Lee, Diana Liskovich, Yinghai Lu,
  Yuning Mao, Xavier Martinet, Todor Mihaylov, Pushkar Mishra, Igor Molybog,
  Yixin Nie, Andrew Poulton, Jeremy Reizenstein, Rashi Rungta, Kalyan Saladi,
  Alan Schelten, Ruan Silva, Eric~Michael Smith, R.~Subramanian, Xia Tan, Binh
  Tang, Ross Taylor, Adina Williams, Jian~Xiang Kuan, Puxin Xu, Zhengxu Yan,
  Iliyan Zarov, Yuchen Zhang, Angela Fan, Melanie Kambadur, Sharan Narang,
  Aurelien Rodriguez, Robert Stojnic, Sergey Edunov, and Thomas Scialom. 2023.
\newblock \href {https://api.semanticscholar.org/CorpusID:259950998} {Llama 2:
  Open foundation and fine-tuned chat models}.
\newblock \emph{ArXiv}, abs/2307.09288.

\bibitem[{Tu et~al.(2016)Tu, Lu, Liu, Liu, and Li}]{tu-etal-2016-modeling}
Zhaopeng Tu, Zhengdong Lu, Yang Liu, Xiaohua Liu, and Hang Li. 2016.
\newblock \href {https://doi.org/10.18653/v1/P16-1008} {Modeling coverage for
  neural machine translation}.
\newblock In \emph{Proceedings of the 54th Annual Meeting of the Association
  for Computational Linguistics (Volume 1: Long Papers)}, pages 76--85, Berlin,
  Germany. Association for Computational Linguistics.

\bibitem[{Williams et~al.(2018)Williams, Nangia, and
  Bowman}]{williams-etal-2018-broad}
Adina Williams, Nikita Nangia, and Samuel Bowman. 2018.
\newblock \href {https://doi.org/10.18653/v1/N18-1101} {A broad-coverage
  challenge corpus for sentence understanding through inference}.
\newblock In \emph{Proceedings of the 2018 Conference of the North {A}merican
  Chapter of the Association for Computational Linguistics: Human Language
  Technologies, Volume 1 (Long Papers)}, pages 1112--1122, New Orleans,
  Louisiana. Association for Computational Linguistics.

\bibitem[{Xue et~al.(2021)Xue, Constant, Roberts, Kale, Al-Rfou, Siddhant,
  Barua, and Raffel}]{xue-etal-2021-mt5}
Linting Xue, Noah Constant, Adam Roberts, Mihir Kale, Rami Al-Rfou, Aditya
  Siddhant, Aditya Barua, and Colin Raffel. 2021.
\newblock \href {https://doi.org/10.18653/v1/2021.naacl-main.41} {m{T}5: A
  massively multilingual pre-trained text-to-text transformer}.
\newblock In \emph{Proceedings of the 2021 Conference of the North American
  Chapter of the Association for Computational Linguistics: Human Language
  Technologies}, pages 483--498, Online. Association for Computational
  Linguistics.

\end{thebibliography}
